\documentclass{article}

\usepackage{arxiv}

\usepackage[utf8]{inputenc} 
\usepackage[T1]{fontenc}    
\usepackage{hyperref}       
\usepackage{url}            
\usepackage{booktabs}       
\usepackage{amsfonts}       
\usepackage{microtype}      
\usepackage{lipsum}
\usepackage{graphicx}
\graphicspath{ {./images/} }

\usepackage{subcaption} 
\usepackage{xspace}
\usepackage{amsmath}
\usepackage{siunitx}
\usepackage{multirow}
\usepackage{adjustbox}

\DeclareSIUnit\bar{bar}

\NewDocumentCommand \suite {m} {\texttt{#1}}

\usepackage{chemformula}

\usepackage{cleveref}

\usepackage{amssymb}

\title{A Kolmogorov-Arnold Surrogate Model for Chemical
Equilibria: Application to Solid Solutions}

\author{
	Leonardo Boledi \textsuperscript{1,*}, \quad Dirk Bosbach \textsuperscript{1}, \quad Jenna Poonoosamy \textsuperscript{1} \vspace*{2mm} \\
	\textsuperscript{1}{Institute of Fusion Energy and Nuclear Waste Management (IFN-2), Forschungszentrum Jülich GmbH} \\
	\textsuperscript{*}{Corresponding author: \texttt{l.boledi@fz-juelich.de}}\\
}

\begin{document}

	\maketitle
	
	\begin{abstract}
		
		The computational cost of geochemical solvers is a challenging matter. For reactive transport simulations, where chemical calculations are performed up to billions of times, it is crucial to reduce the total computational time. Existing publications have explored various machine-learning approaches to determine the most effective data-driven surrogate model. In particular, multilayer perceptrons are widely employed due to their ability to recognize nonlinear relationships. In this work, we focus on the recent Kolmogorov-Arnold networks, where learnable spline-based functions replace classical fixed activation functions. This architecture has achieved higher accuracy with fewer trainable parameters and has become increasingly popular for solving partial differential equations. First, we train a surrogate model based on an existing cement system benchmark. Then, we move to an application case for the geological disposal of nuclear waste, i.e., the determination of radionuclide-bearing solids solubilities. To the best of our knowledge, this work is the first to investigate co-precipitation with radionuclide incorporation using data-driven surrogate models, considering increasing levels of thermodynamic complexity from simple mechanical mixtures to non-ideal solid solutions of binary \ch{(Ba,Ra)SO4} and ternary \ch{(Sr,Ba,Ra)SO4} systems. On the cement benchmark, we demonstrate that the Kolmogorov-Arnold architecture outperforms multilayer perceptrons in both absolute and relative error metrics, reducing them by 62\% and 59\%, respectively. On the binary and ternary radium solid solution models, Kolmogorov-Arnold networks maintain median prediction errors near \SI{1e-3}{}. This is the first step toward employing surrogate models to speed up reactive transport simulations and optimize the safety assessment of deep geological waste repositories.
		
	\end{abstract}

	\section{Introduction}

    \label{sec:section1}

    Computational methods have become a fundamental research direction in the geoscience domain. By simulating the interaction between coupled physical, chemical, and biological processes, reactive transport modelling (RTM) enables scientists to address critical challenges across diverse fields \cite{steefel_reactive_2005, steefel_reactive_2019}. In subsurface energy applications, reactive transport models are employed to study \ch{CO2} sequestration, geothermal reservoirs, and enhanced oil and gas recovery \cite{kang_pore_2010, liu_reactive_2019, yapparova_advanced_2019, erol_fluid-co2_2022}. Additional applications include the analysis of subsurface contaminant migration and electrokinetic desalination \cite{visser_trends_2009, leterme_reactive_2014, paz-garcia_modeling_2012}.

    For deep geological repositories of radioactive waste, RTM resolves coupled, nonlinear interactions between transport and geochemical reactions at material interfaces and in host rock, controlling the long-term evolution of pH, porosity, corrosion products, and radionuclide migration under experimentally inaccessible repository conditions \cite{MONTENEGRO2023107018, claret_eurad_2024}. These models rely on numerous parameters and competing reaction pathways. Consequently, global sensitivity analysis is required to identify the processes and parameters that dominate safety-relevant outcomes, thereby providing a quantitative basis for uncertainty propagation and for focusing safety and performance assessment on the most influential mechanisms \cite{SAMPER2025106286}. Comprehensive global sensitivity analyses within the safety assessment framework increasingly require the execution of large numbers of reactive transport simulations. This can only be achieved through a combination of physics-based models and data-driven surrogates \cite{Kolditz2023Digitalisation}.

    Despite recent advancements, computational cost remains a major challenge in RTM, as the coupling of chemical and flow solvers requires solving chemical equilibrium calculations millions to billions of times. Namely, chemical equilibrium and/or kinetic calculations involve iterative methods that are called for each mesh cell and at every time step \cite{leal_overview_2017}. Thus, the chemical solver is the most expensive component of the simulation, and it can require up to 10,000x the computational cost of the fluid counterpart \cite{leal_accelerating_2020}. Consequently, significant research efforts have focused on improving the speed of reactive transport simulations in recent years. De Lucia et al.\ \cite{de_lucia_poet_2021}, for instance, developed a distributed hash table approach to cache the results of the geochemical solver, and integrated it in the \suite{POET} simulator \cite{lubke_fast_2025}. Another notable example is found in the solver \suite{Reaktoro}, which employs on-demand machine learning (ML) to extrapolate new chemical states \cite{leal_accelerating_2020}. Within this framework, a speed-up of one to three orders of magnitude has been achieved on flows in heterogeneous porous media \cite{kyas_accelerated_2022}. 

    A different research direction consists in leveraging data-driven methods to fully replace the geochemical solver with a surrogate model. This enables faster prediction of geochemical states without requiring modifications to the underlying solver. Laloy and Jaques developed surrogates based on Gaussian processes, polynomial chaos expansion, and neural networks (NNs) to predict the kinetic sorption of uranium and the leaching of cement matrices \cite{laloy_emulation_2019, laloy_speeding_2022}. Similarly, Demirer et al.\ simulated the hydrothermal dolomitization of a fractured carbonate reservoir via an artificial NN \cite{demirer_improving_2023}, and multiple ML algorithms have been tested on a time-dependent cation exchange problem \cite{silva_rapid_2025}. More comprehensively, Prasianakis et al.\ benchmarked decision trees, Gaussian processes, and NNs against each other for cement hydration and degradation, as well as uranium sorption in claystone systems \cite{prasianakis_geochemistry_2025}. In most of the aforementioned publications, deep NNs exhibit the lowest prediction error, making them a strong candidate for geochemical surrogate models. Given the rapid advancement of deep learning, investigating its latest innovations and their applicability to computational geochemistry is warranted. Among these, Kolmogorov-Arnold networks (KANs) offer a promising approach for complex geochemical systems.

    KANs were first introduced in 2024 \cite{liu_kan_2025}, and have since been employed across multiple fields, including fluid dynamics \cite{guo_physics-informed_2025}, time series analysis \cite{xu_kolmogorov-arnold_2024}, and molecular property prediction \cite{li_kolmogorovarnold_2025}. By replacing classical fixed activation functions with spline-based trainable functions, Liu et al.\ achieved a higher prediction accuracy compared to multilayer perceptrons (MLPs) \cite{liu_kan_2025}. Due to their flexibility, KANs have also been adapted to different NN types, such as convolutional and graph-based architectures \cite{bodner_convolutional_2025, kiamari_gkan_2024}. Despite reports of higher computational times and occasional training instabilities, \cite{shukla_comprehensive_2024, hou_kolmogorov-arnold_2025}, KANs have demonstrated high effectiveness in solving partial differential equations when combined with physics-informed NNs \cite{toscano_pinns_2025, jacob_spikans_2025}. Their ability to capture complex non-linear relationships with few trainable parameters makes them a strong candidate for chemical equilibrium calculations.

    In this contribution, we investigate the applicability of KANs as surrogate models for chemical reactions. First, we consider the hydration and evolution of the cementitious system published in \cite{prasianakis_geochemistry_2025}, and compare the performance of KANs against classical MLPs. We then extend the analysis to sulfate precipitation with radium incorporation, addressing increasing levels of thermodynamic complexity from simple mechanical mixtures to non-ideal solid solutions (SSs), including the binary \ch{(Ba,Ra)SO4} and ternary \ch{(Sr,Ba,Ra)SO4} systems. In particular, we show the accuracy of the KAN surrogate and its computational speedup with respect to the reference solver \suite{GEM-Selektor} \cite{wagner_gems_2012, kulik_gem-selektor_2013}.

    This paper is structured as follows: \Cref{sec:section2} introduces the chemical systems and the data-generation process with \suite{GEM-Selektor}. The computational methods to create the surrogate model, including the MLP and the KAN, are discussed in \Cref{sec:section3}. The performance of the KAN surrogate is analyzed for the cement dissolution benchmark in \Cref{sec:section4}. Then, we show three cases of radium sulfate precipitation up to the ternary non-ideal SS model. In \Cref{sec:section5}, we summarize our findings and discuss potential future directions.

    \section{Chemical Equilibrium Model and Solver}
    \label{sec:section2}

    We now discuss the application case for our surrogate model and the solver employed to generate the training data. Our study investigates the geochemical modelling of radium (Ra) incorporation into sulfate SSs. $^{226}$Ra is a safety-relevant radionuclide for the disposal of spent nuclear fuel and will dominate the main radioactive dose contributor after about 100,000 years \cite{zhang_co-precipitation_2014, skb_report_2022}. Its retention can be achieved through co-precipitation into sulfate. The Ra–sulfate system was selected for our investigation because its thermodynamic properties and solid-solution mixing behavior are well constrained by a substantial body of experimental and numerical work spanning a wide range of temperatures and ionic strengths \cite{Brandt20151, Brandt2018, Brandt20201, Klinkenberg20146620, Klinkenberg20181, Poonoosamy2024Radiochemical, Weber2017722, Vinograd2013398, Vinograd2018190, Vinograd201859}. In particular, we evaluate the aqueous Ra concentrations predicted from solid-solution thermodynamics by systematically increasing the complexity of the mixing model. This is achieved by considering three conceptual representations of Ra incorporation, ranging from (i) idealized mechanical mixtures to (ii) non-ideal binary and (iii) ternary SSs.

    \subsection{Solid Solution Models}

    In case study (i), Ra incorporation is described using a hypothetical SS treated as a mechanical mixture of pure sulfate end members. In this representation, the Gibbs energy of the solid phase $\Delta G^{MM}$ is defined as the mole-fraction-weighted sum of the standard Gibbs energies of formation of the end members, namely
    \begin{equation}
        \Delta G^{MM} = \sum_{i=1}^{n} X_i \, G_i^{\circ}.
        \label{eq:GMM}
    \end{equation}
    Here, $X_i$ is the mole fraction of the $i$-th end member ($i = \ch{BaSO4}$ or \ch{RaSO4}). This formulation represents the limiting case of non-interacting solids and is thermodynamically equivalent to the coexistence of pure phases.

    In case study (ii), the real mixing behavior of \ch{BaSO4} and \ch{RaSO4} is considered and described as a non-ideal regular SS, following \cite{Vinograd201859}. The total Gibbs energy of the SS is expressed as
    \begin{equation}
        \Delta G_{\text{total}}^{\text{real}} = \Delta G^{MM} + \Delta G_{\text{mix}}^{id} + \Delta G^{ex},
        \label{eq:Gtotal}
    \end{equation}
    that is the sum of the mechanical mixture term, the ideal Gibbs energy of mixing, and the excess Gibbs energy of mixing. The ideal Gibbs energy of mixing is given by
    \begin{equation}
        \Delta G_{\text{mix}}^{id} = R\,T \sum_{i=1}^{n} X_i \ln X_i,
        \label{eq:Gmixid}
    \end{equation}

    where $R$ is the gas constant and $T$ denotes the temperature. Non-ideal interactions are accounted for through the excess Gibbs energy term, which is expressed using the Margules interaction parameters
    \begin{equation}
        \Delta G^{ex} = \sum_{i \neq j}^{n} X_i X_j \, w_{ij}.
    \label{eq:Gex}
    \end{equation}
    The term $w_{ij}$ denotes the binary interaction parameter between the end members $i$ and $j$. The activity coefficients of the end members are calculated using the Thomson--Waldbaum model assuming a regular solution behavior, with a Margules interaction parameter $w_{\ch{RaBa}} = \SI{2470}{\joule\per\mol}$ \cite{Vinograd201859}.

    In case study (iii), the SS model is extended to the ternary \ch{RaSO4-BaSO4-SrSO4} system. Following \cite{Vinograd201859}, ternary mixing is described using a regular solution model with pairwise binary interaction parameters $w_{\ch{RaBa}} = \SI{2470}{\joule\per\mol}$, $w_{\ch{SrRa}} = \SI{1750}{\joule\per\mol}$, and $w_{\ch{SrBa}} = \SI{750}{\joule\per\mol}$. \Cref{eq:GMM,eq:Gmixid} are reformulated to include the third end member, while the excess Gibbs energy term accounts for all binary interactions among the three sulfate components. Solid-solution-acqueous-solution equilibria are modeled using the \suite{GEM-Selektor} software \cite{kulik_gem-selektor_2013}, which incorporates the TSolMod library for the mixing in SSs \cite{wagner_gems_2012} and the PSI--Nagra 12/07 chemical thermodynamic database \cite{hummel2002nagra, thoenen2014psi}.

    \subsection{Solver Setup}

    In \suite{GEM-Selektor}, phase equilibria are determined by direct minimization of the total Gibbs energy of the system, which is defined by its bulk elemental composition, temperature, pressure, standard molar Gibbs energies of all chemical species, and mixing parameters for solution phases.
    Thermodynamic data for solid sulfates are taken from Table 2 in \cite{Vinograd201859}, while data for aqueous species are obtained from the PSI--Nagra database \cite{thoenen2014psi}. This database incorporates temperature and pressure dependencies derived primarily from the Helgeson--Kirkham--Flowers equation of state for aqueous ions and complexes, as well as for the water solvent, as implemented in the SUPCRT98 database \cite{Helgeson19811249}. Mixing in the ternary SS is described using the built-in ternary regular solution model of \cite{wagner_gems_2012}. For aqueous species, an electrolyte ion-association model consistent with the SUPCRT92 database is applied together with the built-in extended Debye--H\"uckel model \cite{Johnson1992899}.

    \begin{table}[h!]
        \centering
        \begin{tabular}{c|c|c|c|c|c}
        \textbf{Model} & \textbf{\ch{BaSO4}} (\si{\micro\mol}) & \textbf{\ch{NaCl}} (\si{\milli\mol}) & \textbf{\ch{RaBr2}} (\si{\micro\mol}) & \textbf{\ch{SrSO4}} (\si{\milli\mol}) & \textbf{\textit{T}} ($^\circ$C) \\
        \hline
        (i) & 50 -- 500 & 50 -- 500 & 50 -- 500 & - & 25.0 \\
        (ii) & 50 -- 500 & 50 & 50 -- 500 & - & 20.0 -- 90.0 \\
        (iii) & 50 -- 500 & 50 -- 500 & 50 -- 500 & 5 -- 50 & 25.0 \\
        \end{tabular}
        \caption{Input variables for the \suite{GEM-Selektor} simulations. Note that we consider different temperature values only in case (ii).}
        \label{table:solverSetup}
    \end{table}

    The input configuration for the geochemical solver consists of \SI{1}{\kg} of water and $\SI{1}{\mol}$ of atmospheric air, referred to as "AtmAirNit" in \suite{GEM-Selektor}. Calculations are performed at $P = \SI{1}{\bar}$. \Cref{table:solverSetup} shows the range of the remaining variables for each SS model. The selected values are inspired by the radium incorporation simulation in \cite{wang_contrasting_2026}. Over this range, we select points via Sobol sampling to generate the datasets for the upcoming surrogate models \cite{sobol_distribution_1967}. The thermodynamic dataset for the \ch{(Ba,Sr,Ra)SO4-H2O} system, including the temperature dependence of \ch{RaSO4} solubility, is taken from \cite{Vinograd2018190}.

    \section{Surrogate Models}
    \label{sec:section3}
    
    In this section, we introduce the ML models used to accelerate the chemical equilibrium solver. Specifically, we employ NNs to predict the outcome of the chemical reaction, i.e., output concentrations and pH, based on the system's initial conditions. We emphasise that this approach generates a mapping between input and output data without requiring knowledge of the underlying physics, which we refer to as a surrogate model \cite{williams_novel_2021}. This strategy is different from, e.g., the aforementioned approach from Leal et al.\ \cite{leal_accelerating_2020}, which leverages previous reaction outcomes to speed up the upcoming computations. Once the surrogate model is built, it is meant to fully replace the original solver. We now describe the two main algorithms of this study, MLPs and KANs.

    \subsection{Multilayer Perceptrons}
    
    \begin{figure}
        \centering
        \includegraphics[width=0.49\textwidth]{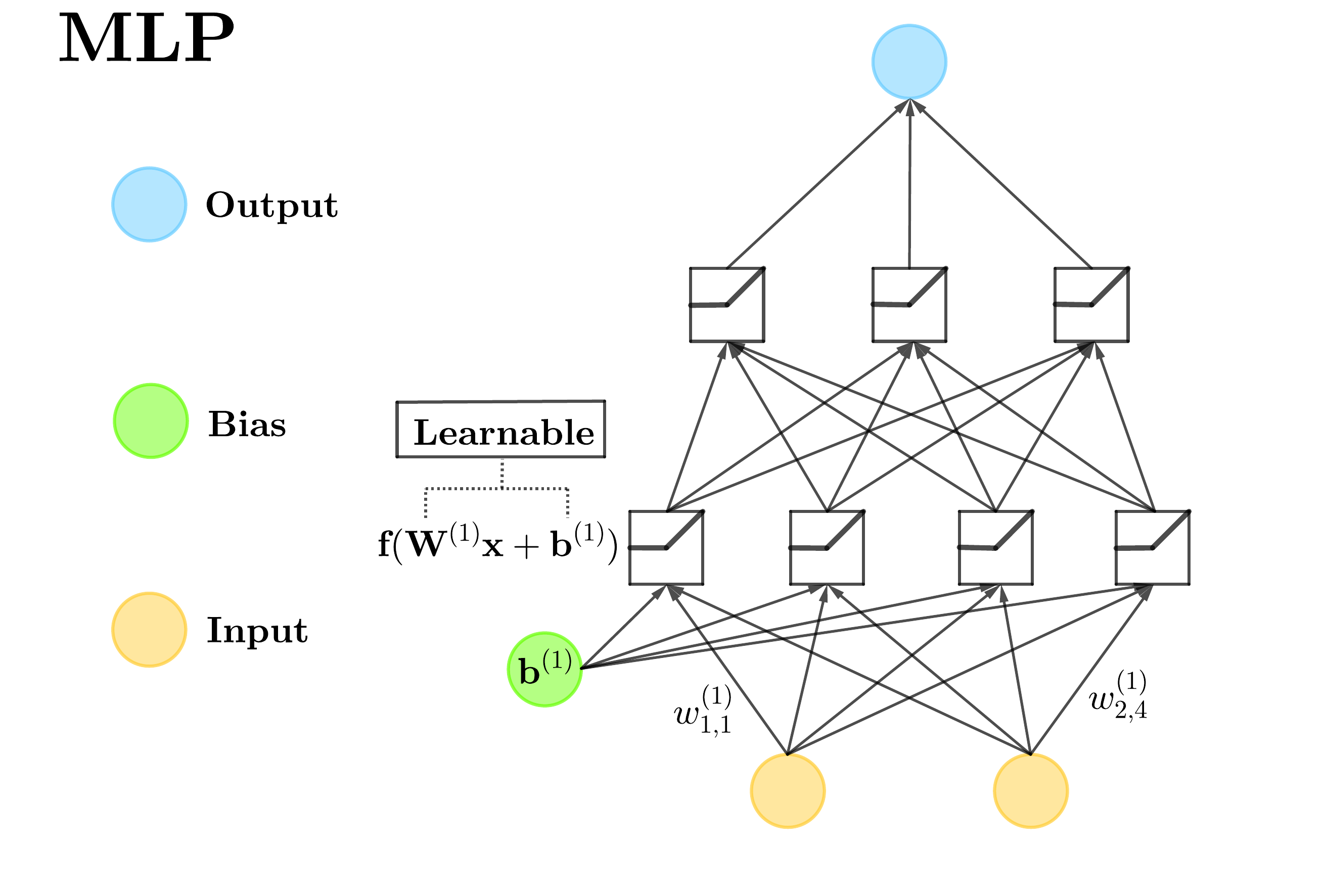}
        \includegraphics[width=0.49\textwidth]{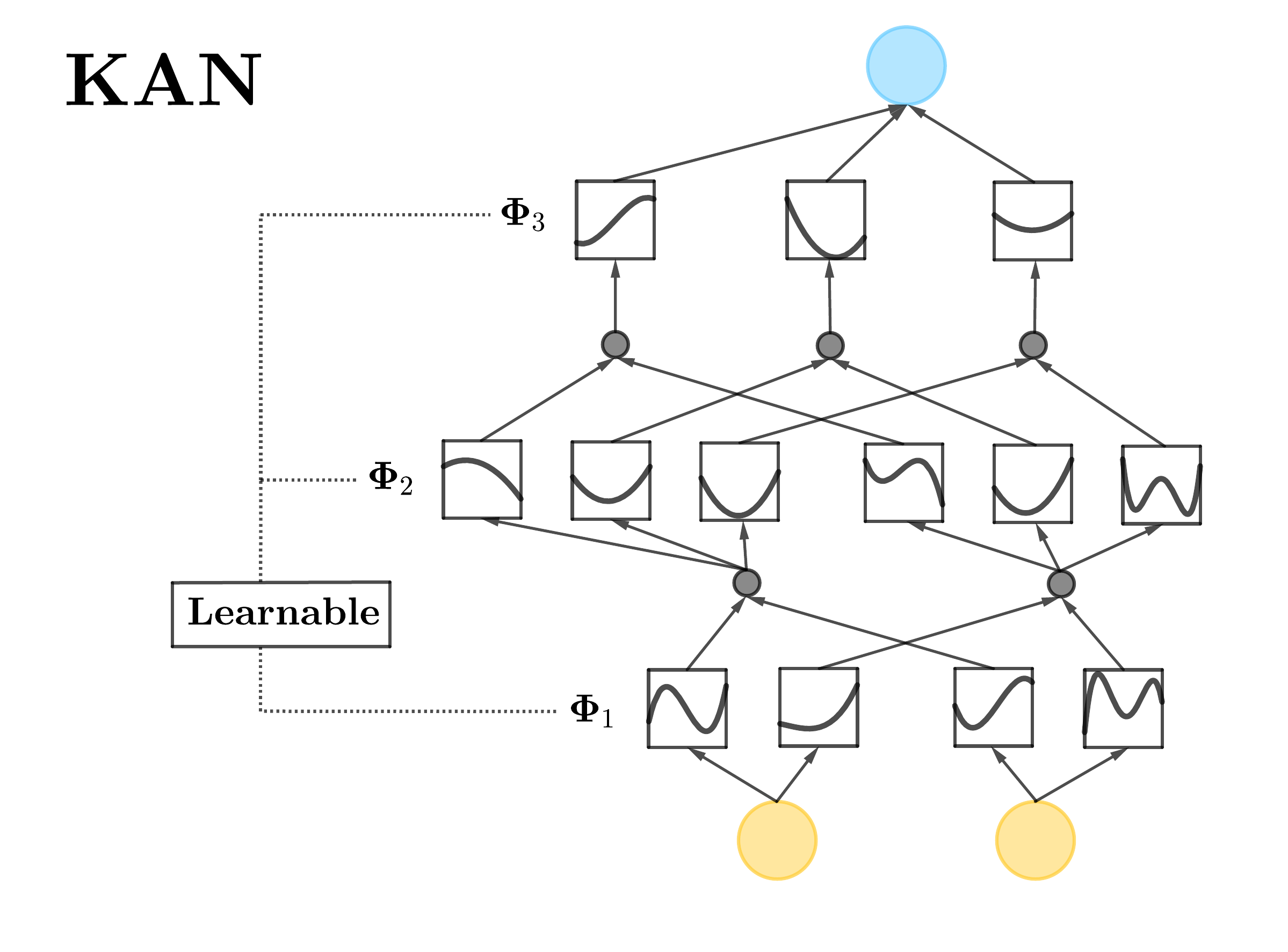}
        \caption{Sketch of an MLP (left) and a KAN (right) with two hidden layers. In KANs, instead of training for weights and biases, learnable activation functions are placed on each edge.}
        \label{fig:nnSketches}
    \end{figure}
    
    Let us consider a vector of observations $\mathbf{x}\in\mathcal{X}\subseteq\mathbb{R}_+^p$, with $p\in\mathbb{N}$, which represents the initial state of a chemical system. Let $\mathbf{y}\in\mathcal{Y}\subseteq\mathbb{R}_+^q$, with $q\in\mathbb{N}$, be the vector of outputs, such as the equilibrium concentration and speciation. An NN is a mathematical
    function $\mathbf{F}_{\boldsymbol{\theta}} : \mathcal{X}\rightarrow\mathcal{Y}$ depending on a set of parameters $\boldsymbol{\theta} \in \boldsymbol{\Theta}$ that associates to any observation point $\mathbf{x}$ the output point $\mathbf{y}$. It comprises computational units, the so-called neurons, which take the form
    \begin{equation}
        h(\textbf{x}) = f(\textbf{w}^\top\textbf{x} + b),
        \label{eq:nn_neuron}
    \end{equation}
    where $f$ is the real-valued activation function, $\textbf{w}$ is the vector of weights, and $b$ represents the bias associated with the neuron. During the training process, the weights and biases associated with each neuron are selected to minimize the prediction error.
    
    An MLP is a feedforward fully-connected NN consisting of at least one hidden layer \cite{prasianakis_geochemistry_2025}. That is, an NN without directed loops or cycles across neurons and where every neuron of a given layer is connected to all the neurons of the previous layer, see \Cref{fig:nnSketches}. Mathematically, the extension of \Cref{eq:nn_neuron} to an MLP of $n$ hidden layers can be expressed as a nested composition of affine transformations and activation functions, which takes the form
    \begin{eqnarray}
        \textbf{y} = \textbf{F}_{\boldsymbol{\theta}}(\textbf{x}) = \textbf{f}_{n+1}\Big(\textbf{W}^{(n+1)}\textbf{f}_n\Big(\textbf{W}^{(n)}\textbf{f}_{n-1}\Big(\textbf{W}^{(n-1)}\dots\textbf{f}_2\Big(\textbf{W}^{(2)}\textbf{f}_1\Big(\textbf{W}^{(1)}\textbf{x} + \textbf{b}^{(1)}\Big) + \textbf{b}^{(2)}\Big) \nonumber \\ + \dots \textbf{b}^{(n-1)}\Big) + \textbf{b}^{(n)}\Big) + \textbf{b}^{(n+1)}\Big).
        \label{eq:mlp}
    \end{eqnarray}
    Here, $\textbf{f}_{i}$ represent the vector of real valued functions from $\mathbb{R}^{c_h}$ to $\mathbb{R}^{c_h}$ for $i=1,\dots,n+1$, $\textbf{W}^{(i)}$ is the weight matrix, and $c_h$ denotes the number of neurons in the $h$-th hidden layer. A more detailed analysis of the MLP architecture is beyond the scope of this work and can be found, e.g., in \cite{hounmenou_formalism_2021, chan_deep_2023}.
    
    The network is implemented and trained in the \suite{Pytorch Lightning} framework, which allows straightforward extension to GPU acceleration \cite{Falcon_PyTorch_Lightning_2019}. For the selection of hyperparameters such as the number of layers, neurons, and the batch size, we employ the Bayesian optimiser \suite{Optuna} \cite{akiba_optuna_2019}.
    
    \subsection{Kolmogorov-Arnold Networks}
    
    The main idea behind the KAN lies in the Kolmogorov-Arnold representation theorem, which states that every multivariate continuous function can be written as a finite composition of continuous functions of a single variable and the binary operation of addition \cite{liu_kan_2025}. Based on this, Liu et al.\ proposed the alternative neuron formulation
    \begin{equation}
        \phi(x) = w_f f(x) + w_s\sum_j c_j B_{j,d}(x),
        \label{eq:kan_neuron}
    \end{equation}
    where $x$ represents a scalar input to simplify the notation, and $w_f$ and $w_s$ are learnable scaling parameters. The first term denotes a classical fixed activation function (commonly the SiLU activation), while the second term is a learnable spline component. Specifically, $B_{j,d}(x)$ denotes the
    $j$-th B-spline basis function of degree
    $d$ \cite{prautzsch_bezier_2002}, and $c_j$ are the learnable spline coefficients optimised during training to minimise prediction error.

    The extension of \Cref{eq:kan_neuron} to a KAN of $n$ hidden layers can be expressed as the composition
    \begin{equation}
        \textbf{f}(\textbf{x}) = (\boldsymbol{\Phi}_{n+1} \circ \boldsymbol{\Phi}_{n} \circ \dots \circ \boldsymbol{\Phi}_{2} \circ \boldsymbol{\Phi}_{1})(\textbf{x}),
        \label{eq:kan_formulation}
    \end{equation}
    where $\boldsymbol{\Phi}_i$ denotes the $i$-th KAN layer. Here, $\boldsymbol{\Phi}_i = \{\phi_{i,p,q}\}$ represents the matrix of 1D activation functions connecting input neuron $p$ to output neuron $q$ in layer $i$, as defined in \cite{liu_kan_2025}. For further details, the reader is referred to the original publication. 
    
    A key distinction from MLPs is that KANs place learnable activation functions on the edges between neurons rather than using fixed activations at the nodes, see \Cref{fig:nnSketches} for a visual comparison. The degree $d$ and number of control points for these spline-based activations can be flexibly chosen by the user. This flexibility enables KANs to achieve lower prediction errors compared to MLPs with fewer trainable parameters \cite{liu_kan_2025}.
    
    For the implementation and training, we leverage the \suite{efficient-kan} package \cite{Blealtan_efficient_kan_2024}, which offers a KAN model compatible with \suite{Pytorch Lightning}. Hyperparameters, including the number of layers, neurons per layer, and spline degree, are selected using the Bayesian optimisation library \suite{Optuna} \cite{akiba_optuna_2019}.

    \section{Results}
    \label{sec:section4}
    
    After having introduced the surrogate model, we demonstrate the capability of the KAN across four test cases. We start by examining the cement hydration benchmark, then we move to radium uptake into SSs containing barium and strontium.
    
    \subsection{Cement System \ch{CaO-SiO2-H2O}}
    \label{sec:cemResults}
    
    To assess the performance of the KAN against classical approaches, we first consider the benchmark published in \cite{prasianakis_geochemistry_2025}. Here, the authors examined the hydration of the \ch{CaO-SiO2-H2O} cementitious system under isothermal conditions at 25 °C. The input is 3-dimensional (\ch{CaO}, \ch{SiO2}, and \ch{H2O}), and the output is 18-dimensional. The output covers the aqueous species after equilibration, as well as the amounts and composition of the solid phases and SSs.
    
    As the training data is published \cite{prasianakis_repo_2025}, we leverage the 50,000 sample points in \path{10_PC_02_LHS_50000_54854_01_s1_G.csv} for our dataset. We refer to \cite{prasianakis_geochemistry_2025} for details on the employed geochemical solver and the data generation process. To train the NNs, we randomly select 40,000 points from the dataset, while the remaining part is split in half for the validation and test phases, respectively. As preprocessing, log transformation and min-max scaling are applied to both the input and output data. In their study, Prasianakis et al.\ trained a 5-layer MLP with 192 neurons per layer and Mish activation functions, which we keep as the reference model. For the KANs, we consider two models of different sizes. The first network consists of 4 hidden layers with 28 neurons per layer, 7th-order activation functions, and 10 grid points, resulting in approximately one-third of the trainable parameters compared to the MLP model. The second KAN consists of 5 layers with 40 neurons per layer, 8-th order activation functions, and 12 grid points. All networks are trained over 200 epochs, with a batch size of 192 and the mean squared error loss function. Finally, a learning rate scheduler is employed to avoid local minima, starting with a value of 0.01. If the validation error does not improve in 10 iterations, the learning rate is reduced by $90\%$.
    
    \begin{figure}
        \centering
        \includegraphics[width=\textwidth]{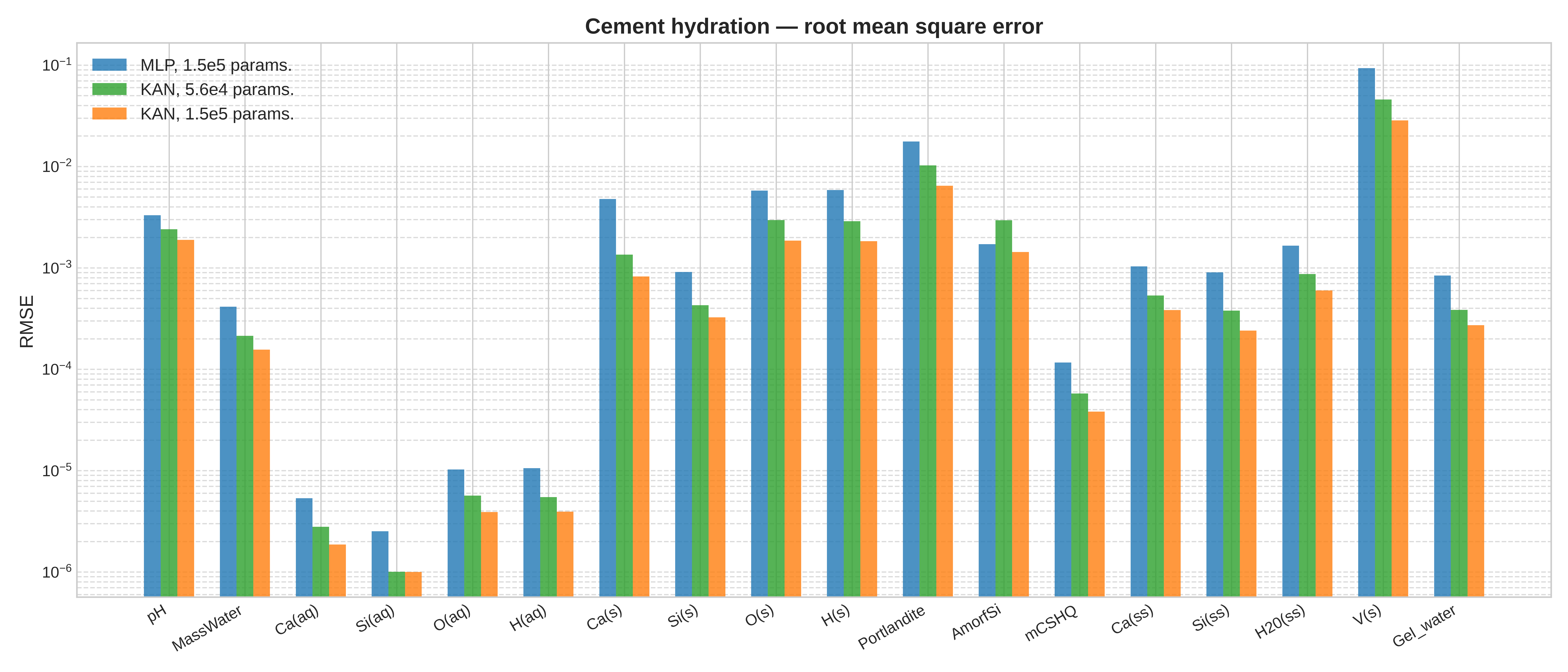}
        \caption{RMSE on the test set for the cement hydration case. The reference MLP model (blue) is shown against two KANs of different sizes (green and orange).}
        \label{fig:cementRMSE}
    \end{figure}
    
    \begin{table}[ht]
        \centering
        \begin{tabular}{c|c|c|c}
            \textbf{Variable} & \textbf{RMSE\textsubscript{MLP}} & \textbf{RMSE\textsubscript{KAN}} & \textbf{Improvement (\%)} \\
            \hline
            pH  & $3.3071\times10^{-3}$ & $1.8907\times10^{-3}$ & 42.83 \\
            MassWater  & $4.1398\times10^{-4}$ & $1.5640\times10^{-4}$ & 62.22 \\
            Ca(aq)  & $5.3386\times10^{-6}$ & $1.8701\times10^{-6}$ & 64.97 \\
            Si(aq)  & $2.5183\times10^{-6}$ & $1.0015\times10^{-6}$ & 60.23 \\
            O(aq)  & $1.0262\times10^{-5}$ & $3.9045\times10^{-6}$ & 61.95 \\
            H(aq)  & $1.0571\times10^{-5}$ & $3.9398\times10^{-6}$ & 62.73 \\
            Ca(s)  & $4.7758\times10^{-3}$ & $8.2431\times10^{-4}$ & 82.74 \\
            Si(s)  & $9.1093\times10^{-4}$ & $3.2470\times10^{-4}$ & 64.35 \\
            O(s)  & $5.7943\times10^{-3}$ & $1.8594\times10^{-3}$ & 67.91 \\
            H(s) & $5.8698\times10^{-3}$ & $1.8339\times10^{-3}$ & 68.76 \\
            Portlandite & $1.7666\times10^{-2}$ & $6.4487\times10^{-3}$ & 63.50 \\
            AmorfSi & $1.7160\times10^{-3}$ & $1.4347\times10^{-3}$ & 16.39 \\
            mCSHQ & $1.1664\times10^{-4}$ & $3.8269\times10^{-5}$ & 67.19 \\
            Ca(ss) & $1.0327\times10^{-3}$ & $3.8317\times10^{-4}$ & 62.90 \\
            Si(ss) & $9.0186\times10^{-4}$ & $2.4104\times10^{-4}$ & 73.27 \\
            \ch{H_2O}(ss) & $1.6592\times10^{-3}$ & $5.9939\times10^{-4}$ & 63.88 \\
            V(s) & $9.3427\times10^{-2}$ & $2.8549\times10^{-2}$ & 69.44 \\
            Gel\_water & $8.3947\times10^{-4}$ & $2.7253\times10^{-4}$ & 67.54 \\
        \end{tabular}
        \caption{Prediction errors on the test set for the cement hydration test case. The MLP and KAN models with \SI{1.5e5}{} parameters are shown, together with the percentage improvement of the RMSE.}
        \label{table:cementRMSE}
    \end{table}
    
    \Cref{fig:cementRMSE} shows the root mean squared error (RMSE) on each output variable. The blue bars denote the MLP model according to \cite{prasianakis_geochemistry_2025}, while the green and orange bars represent the KAN. Note that we cannot retrieve the same error value as in \cite{prasianakis_geochemistry_2025} since the training process is not fully reproducible, but the overall behavior is comparable, as the formation of minerals and the total solid volume are once again the most challenging quantities to predict. In terms of absolute error, the KANs outperform the MLP model across all output variables, except for amorphous silica in the smaller network. In particular, if we pick a KAN with the same number of trainable parameters, an average error improvement of $62\%$ is achieved, see \Cref{table:cementRMSE}.
    
    \begin{figure}
        \centering
        \includegraphics[width=\textwidth]{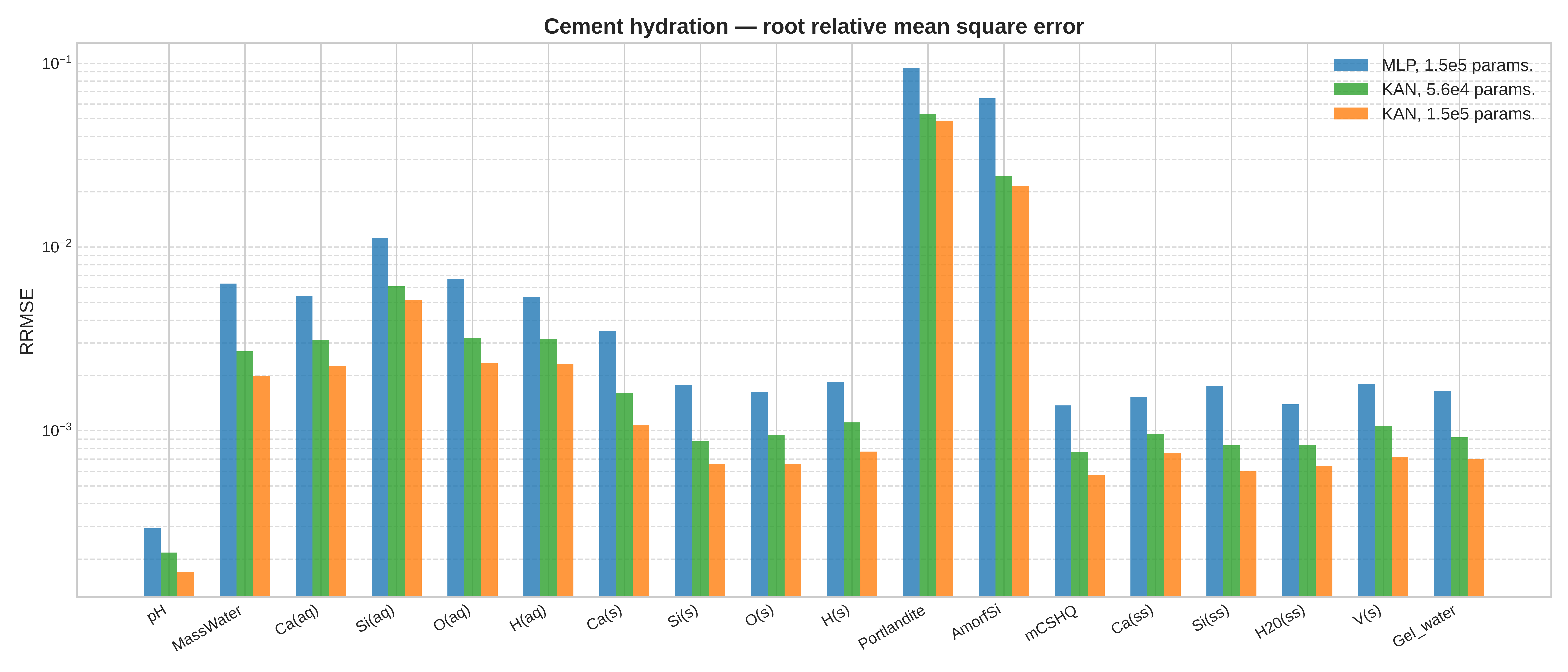}
        \caption{RRMSE on the test set for the cement hydration case. The reference MLP model (blue) is shown against two KANs of different sizes (green and orange).}
        \label{fig:cementRRMSE}
    \end{figure}
    
    \begin{table}[ht]
        \centering
        \begin{tabular}{c|c|c|c}
        \textbf{Variable} & \textbf{RRMSE\textsubscript{MLP}} & \textbf{RRMSE\textsubscript{KAN}} & \textbf{Improvement (\%)} \\
        \hline
        pH           & $2.9427\times10^{-4}$  & $1.7009\times10^{-4}$  & 42.20 \\
        MassWater    & $6.3279\times10^{-3}$ & $1.9846\times10^{-3}$ & 68.64 \\
        Ca(aq)       & $5.4218\times10^{-3}$ & $2.2426\times10^{-3}$ & 58.64 \\
        Si(aq)       & $1.1215\times10^{-2}$ & $5.1705\times10^{-3}$ & 53.90 \\
        O(aq)        & $6.7081\times10^{-3}$ & $2.3244\times10^{-3}$ & 65.35 \\
        H(aq)        & $5.3405\times10^{-3}$ & $2.3007\times10^{-3}$ & 56.92 \\
        Ca(s)        & $3.4789\times10^{-3}$ & $1.0662\times10^{-3}$ & 69.35 \\
        Si(s)        & $1.7738\times10^{-3}$ & $6.6102\times10^{-4}$  & 62.74 \\
        O(s)         & $1.6295\times10^{-3}$ & $6.6046\times10^{-4}$  & 59.47 \\
        H(s)         & $1.8466\times10^{-3}$ & $7.6938\times10^{-4}$  & 58.34 \\
        Portlandite  & $9.4112\times10^{-2}$ & $4.8741\times10^{-2}$ & 48.21 \\
        AmorfSi      & $6.4440\times10^{-2}$ & $2.1477\times10^{-2}$ & 66.67 \\
        mCSHQ        & $1.3699\times10^{-3}$ & $5.7155\times10^{-4}$  & 58.28 \\
        Ca(ss)       & $1.5273\times10^{-3}$ & $7.5066\times10^{-4}$  & 50.85 \\
        Si(ss)       & $1.7567\times10^{-3}$ & $6.0620\times10^{-4}$ & 65.49 \\
        \ch{H_2O}(ss) & $1.3894\times10^{-3}$ & $6.4193\times10^{-4}$ & 53.80 \\
        V(s)         & $1.7991\times10^{-3}$ & $7.2056\times10^{-4}$  & 59.95 \\
        Gel\_water   & $1.6498\times10^{-3}$ & $6.991\times10^{-4}$  & 57.63 \\
        \end{tabular}
        \caption{Prediction errors on the test set for the cement hydration test case. The MLP and KAN models with \SI{1.5e5}{} parameters are shown, together with the percentage improvement of the RRMSE.}
        \label{table:cementRRMSE}
    \end{table}
    
    When evaluating chemical reactions, we often encounter output variables with different orders of magnitude. Thus, it is paramount to examine the relative error of the predictions. Following \cite{prasianakis_geochemistry_2025}, the relative root mean squared error (RRMSE) is defined as
    \begin{equation}
        \text{RRMSE} = \sqrt{\frac{1}{N}\sum_i(e_i)^2}, \hspace{3mm} \text{with} \hspace{3mm} e_i = \begin{cases} (y_i - \hat{y}_i)/y_i \hspace{3mm} \text{if} \hspace{3mm} y_i,\hat{y}_i \neq 0, \\ 1 \hspace{21mm} \text{if} \hspace{3mm} y_i=0 \text{ and } \hat{y}_i \neq 0, \\ 0 \hspace{21mm} \text{if} \hspace{3mm} y_i=0 \text{ and } \hat{y}_i = 0, \end{cases}
        \label{eq:RRMSE}
    \end{equation}
    where $y_i$ and $\hat{y}_i$ denote the true values and the NN predictions, respectively. Since the formation of minerals, e.g., portlandite, occurs only under certain system conditions, \Cref{eq:RRMSE} avoids division by zero in the error computation. \Cref{fig:cementRRMSE} shows the RRMSE on each output variable. The blue bars denote the MLP model according to \cite{prasianakis_geochemistry_2025}, while the green and orange bars represent the KANs. In terms of relative error, the KANs consistently outperform the MLP model across all output variables, even when selecting the network with fewer trainable parameters. The bigger network achieves an average improvement of $59\%$, as shown in \Cref{table:cementRRMSE}.
    
    \begin{table}[ht]
        \centering
        \scalebox{1.0}{
        \begin{tabular}{c|c|c}
            \textbf{Model} & \textbf{Trainable parameters} & \textbf{Training time (\unit{\second})} \\
            \hline
            MLP & $5.6\times10^4$ & 153 \\
            MLP & $1.5\times10^5$ & 162 \\
            KAN & $5.6\times10^4$ & 600 \\
            KAN & $1.5\times10^5$ & 968
        \end{tabular}}
        \caption{Comparison of training times between KANs and MLPs for the cement hydration test case across different network sizes.}
        \label{table:cementTrainTime}
    \end{table}
    
    As a final test, we examine the training time of both architectures on an NVIDIA Quadro GV100 GPU. \Cref{table:cementTrainTime} shows the training times of the networks discussed above, plus an additional MLP with 4 hidden layers and 132 neurons per layer for comparison. Shukla et al.\  reported significantly slower training times for KANs compared to MLPs \cite{shukla_comprehensive_2024}, attributing this to the backpropagation step becoming less efficient as the complexity of the activation functions increases. Our results confirm their findings: the KAN takes almost four times longer to train for the smallest network configuration, with this gap widening for larger networks due to the increased complexity of learnable spline-based activations. We underline, however, that training is a one-time cost in the modeling workflow (solver setup, data generation, network setup and training, etc.), whereas the higher accuracy achieved by KANs translates to improved predictions across millions or billions of subsequent evaluations in reactive transport simulations.
    
    It is evident then that the KAN architecture is preferable as a surrogate model for the examined cement system. We underline that the objective of this benchmark is to investigate how accurate KANs are against a pre-published NN model. While the average RRMSE remains high when predicting portlandite and amorphous silica, the applicability of this specific surrogate in a real-world scenario is beyond the scope of this work.
    
    \subsection{Radium Uptake (i) - Mechanical Mixing}
    \label{sec:mecResults}
    
    For the upcoming test cases, we move to an application of interest, namely the formation of Ra SSs. We start by considering the simplest system without any ``real'' SS model, i.e., the so-called mechanical mixing. The input is 3-dimensional (\ch{BaSO4}, \ch{NaCl}, and \ch{RaBr2}), and the output is 8-dimensional. Here, we only track the radium species and the pH as quantities of interest.
    
    \begin{figure}
        \centering
        \includegraphics[width=\textwidth]{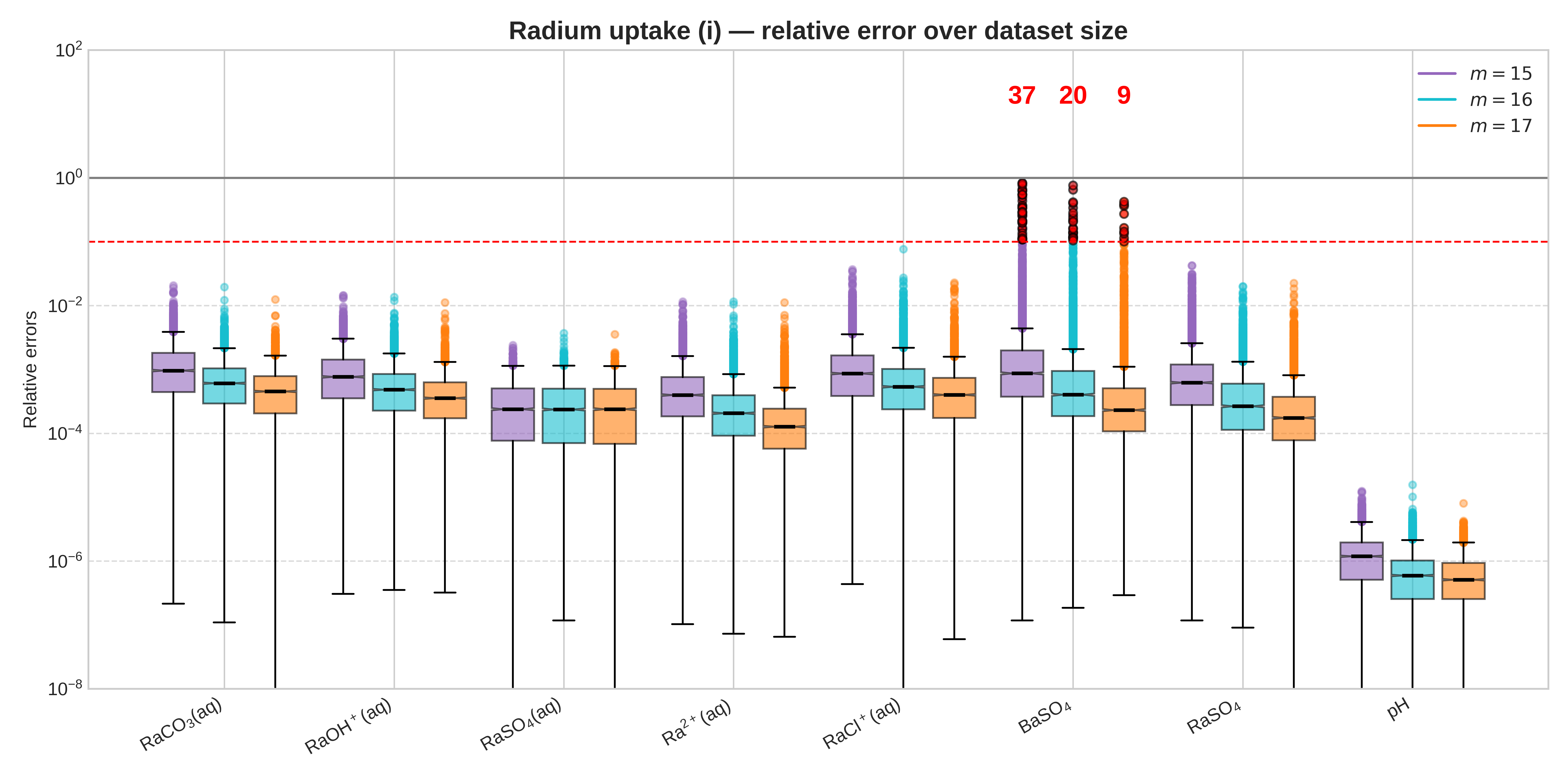}
        \caption{Relative error on the test set for the radium uptake case with mechanical mixing. The errors are plotted for three KANs trained on datasets of different sizes, where $m$ denotes the exponent of the Sobol sampler. In red, we show the number of predictions with an error above $10\%$.}
        \label{fig:mecMixSize}
    \end{figure}
    
    First, we compare the effect size on the surrogate model accuracy. With the solver \suite{GEM-Selektor}, three datasets of size $2^m$ are generated, where $m=15,16,17$. The input data are chosen according to a Sobol sampler, then split in half to obtain the training and validation sets. From the latter, 5000 points are subtracted for testing. This time, the KAN architectures are selected with the hyperparameter-optimization tool \suite{Optuna}. The optimizer finds the best network configuration up to a maximum of 4 hidden layers, 24 neurons per layer, 8th-order activation functions, and 15 grid points. All networks are trained over 100 epochs, with a batch size of 192 and the mean squared error loss function. A learning rate scheduler is employed as in \Cref{sec:cemResults}, but with an 80\% reduction factor.
    
    \Cref{fig:mecMixSize} shows the relative error for all the output variables and dataset sizes. All the error values are combined in a box plot for a more detailed examination. While the median errors are all below \SI{1e-3}, it is also important to check how many predictions are accurate enough for the application of the surrogate model. Note that even with 32,768 ($2^{15}$) data points, only 37 evaluations out of 5000 exceed the $10\%$ error threshold. Additionally, both the median error and the number of high-error points decrease as the size of the employed dataset increases. As training points can be easily generated with \suite{GEM-Selektor}, the model can be refined even further up to the desired accuracy.
    
    \begin{figure}
        \centering
        \includegraphics[width=\textwidth]{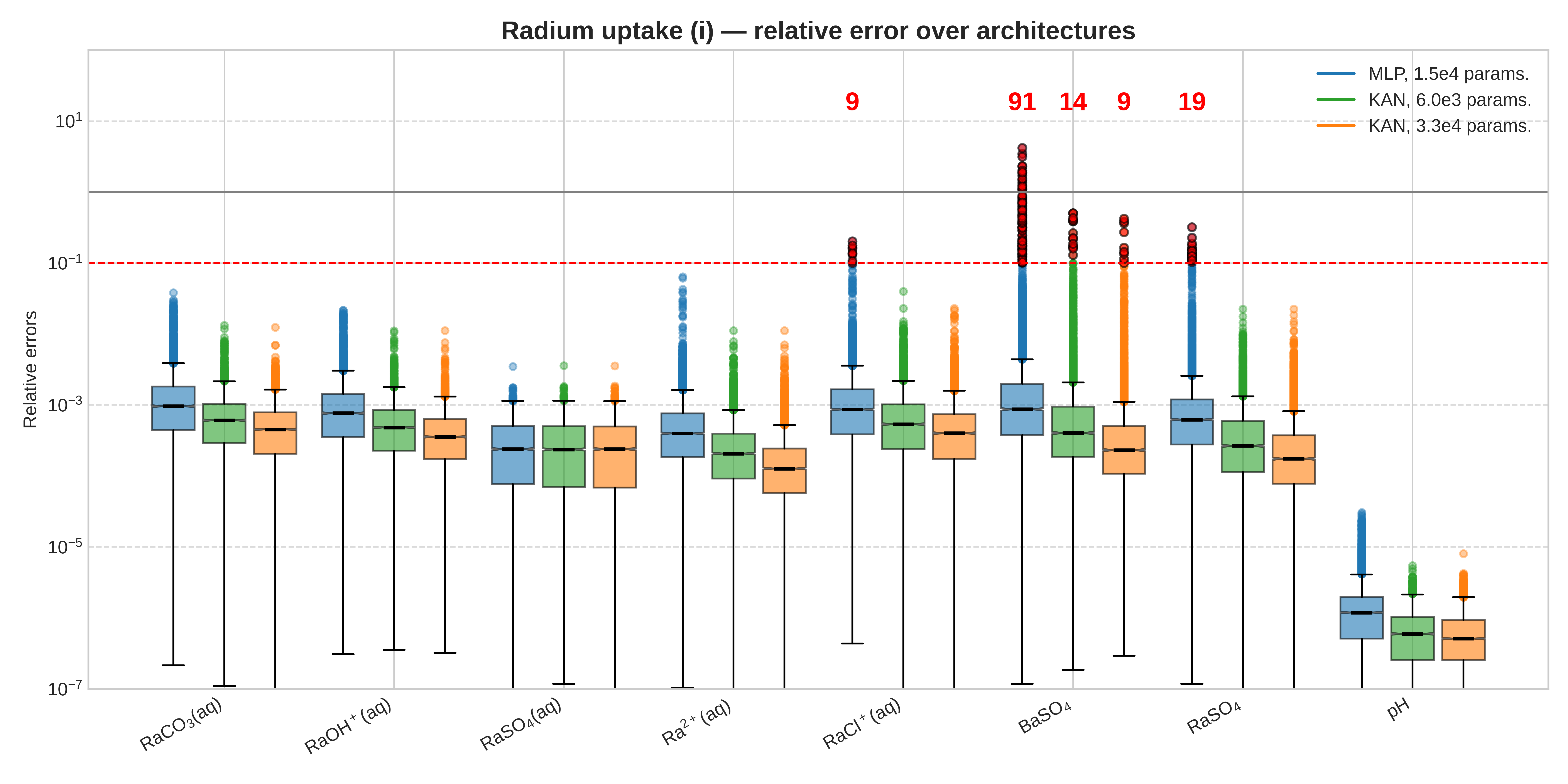}
        \caption{Relative error on the test set for the radium uptake case with mechanical mixing. The classical MLP model (blue) is shown against two KANs of different sizes (green and orange). In red, we show the number of predictions with an error above $10\%$.}
        \label{fig:mecMixErr}
    \end{figure}
    
    We now keep the KAN model trained on the 131,072 ($2^{17}$) points dataset as a reference and compare it once again with an MLP. The hyperparameters are selected with \suite{Optuna} up to a maximum of 10 layers and 64 neurons per layer. Additionally, a KAN with less than half of the trainable parameters is considered. \Cref{fig:mecMixErr} shows the relative error for all the output variables on the different NNs. Once again, both KANs exhibit a lower median error on all the output variables (green and orange), even when considering the small-size network. We underline that the KANs show fewer high-error predictions, as low as one tenth, and only for one single output. As in the cement hydration test case, the KAN architecture performs best.
    
    \subsection{Radium Uptake (ii) - \ch{(Ba,Ra)SO4} Solid Solution}
    
    \begin{figure}
        \centering
        \includegraphics[width=\textwidth]{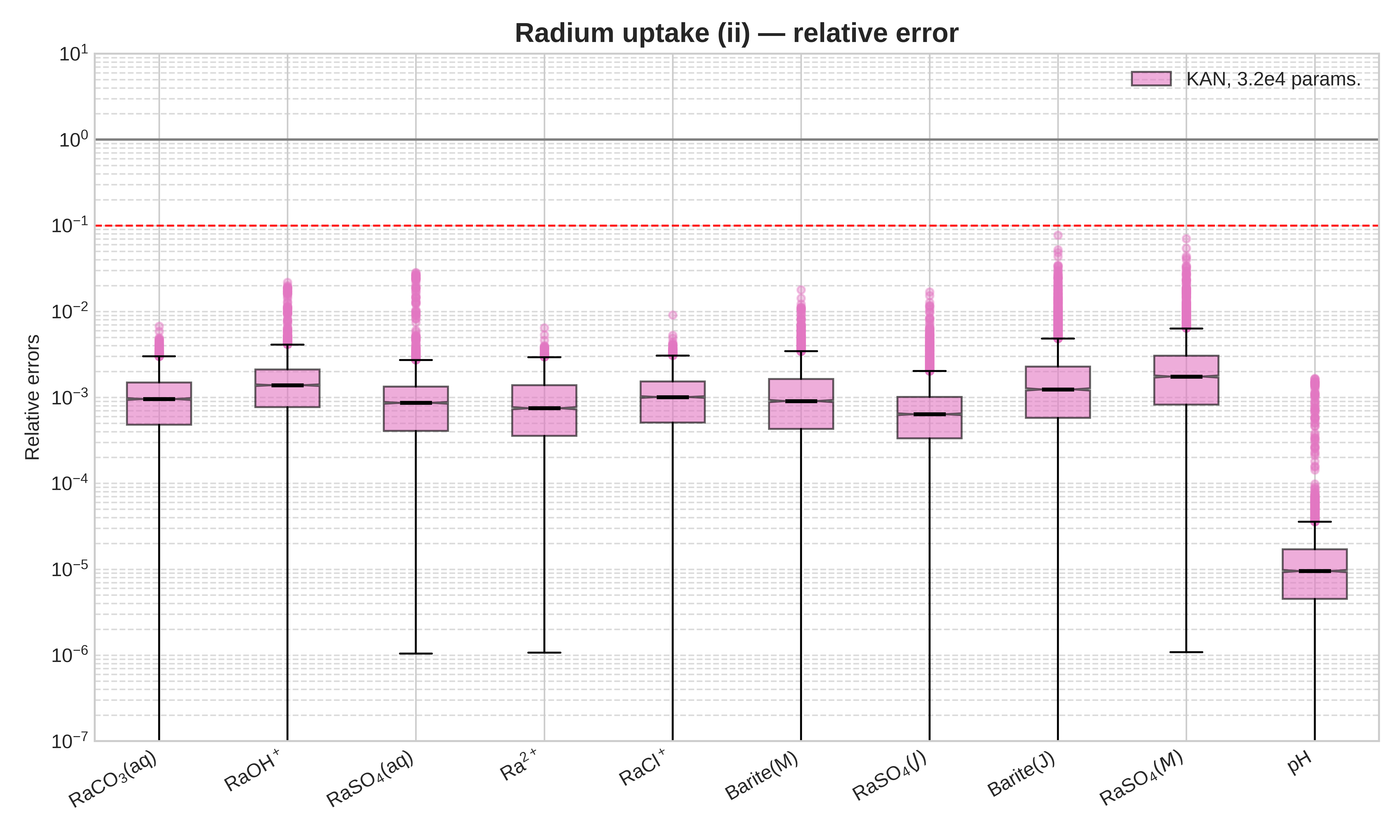}
        \caption{Relative error on the test set for the radium uptake case with the two-component SS model.}
        \label{fig:ss2Comp}
    \end{figure}
    
    After having proven the effectiveness of KANs on the reference problems, we move towards more complex chemical systems by considering SS models. In this example, the two-component \ch{(Ba,Ra)SO4} SS is examined. The input is 3-dimensional (\ch{BaSO4}, \ch{RaBr2}, and $T$) and the output is 10-dimensional, where $T$ is the system's temperature. Again, we only track the radium species and the pH as quantities of interest.
    
    A dataset of 262,144 ($2^{18}$) points is generated using the \suite{GEM-Selektor} solver. The dataset splitting process follows as in \Cref{sec:mecResults}. A KAN with three hidden layers, 24 neurons per layer, 8th-order activation functions, and 12 grid points is selected, resulting in ca.\ 32,200 trainable parameters. The network is trained over 100 epochs, with a batch size of 192 and the mean squared error loss function.
    
    \Cref{fig:ss2Comp} shows the relative errors for all the output variables. Note that the median errors are all below the value \SI{2e-3}{} and no predictions are above the 10\% error threshold. Thus, the KAN proves to be an effective surrogate model for the two-component SS as well. We underline that, for the first time, temperature dependency is considered as a variable in the surrogate model, in contrast to recent publications such as \cite{peng_machine_2025}.
    
    \subsection{Radium Uptake (iii) - \ch{(Sr,Ba,Ra)SO4} Solid Solution}
    
    \begin{figure}
        \centering
        \includegraphics[width=\textwidth]{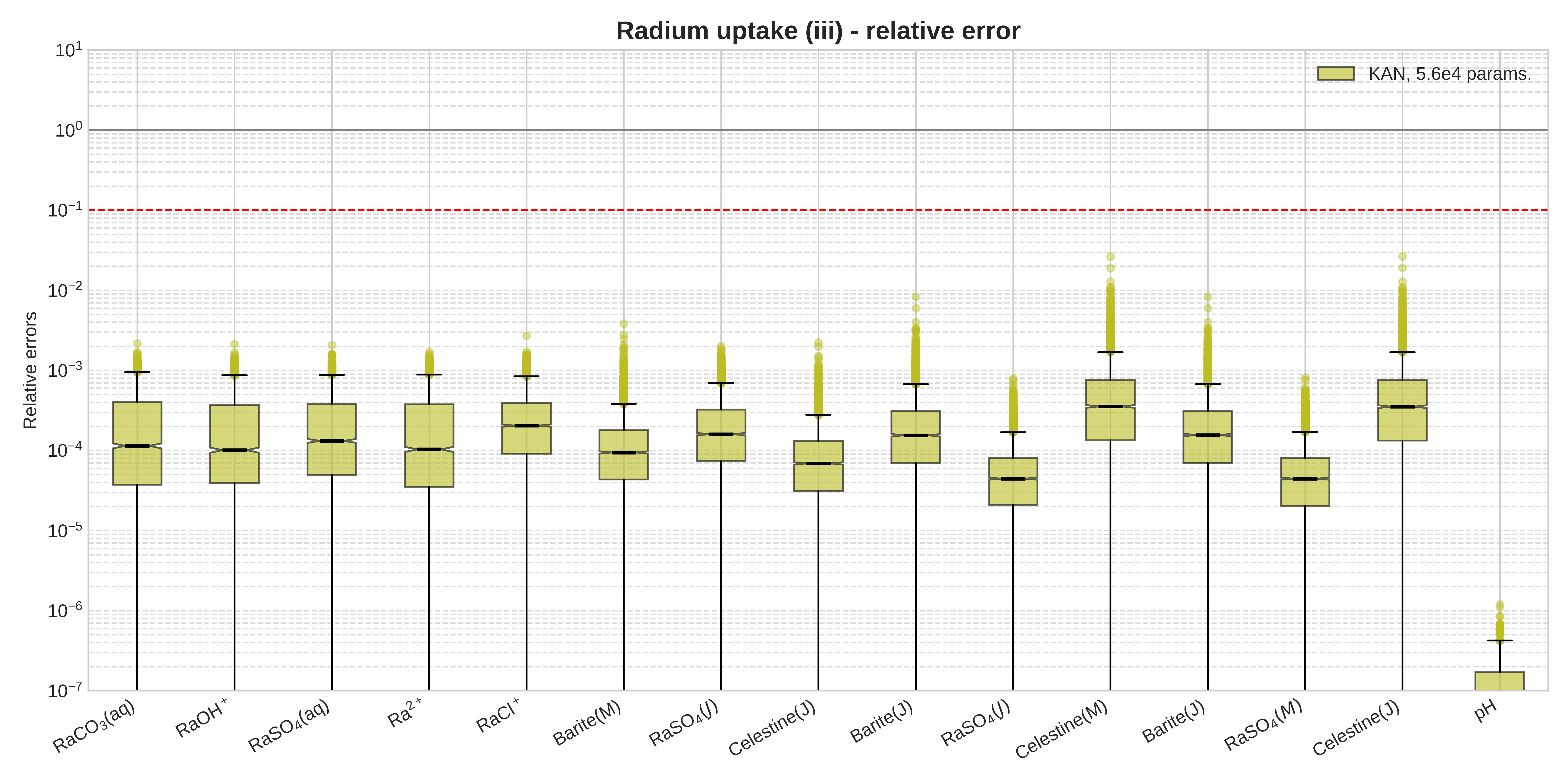}
        \caption{Relative error on the test set for the radium uptake case with the three-component SS model.}
        \label{fig:ss3Comp}
    \end{figure}
    
    For the most complex test case, we examine the ternary \ch{(Sr,Ba,Ra)SO4} SS model with the addition of strontium. The input is four-dimensional (\ch{BaSO4}, \ch{NaCl}, \ch{RaBr2}, \ch{SrSO4}) and the output is 15-dimensional. Only the radium-containing species and the pH are tracked.
    
    We prepare a dataset of 524,000 ($2^{19}$) points using \suite{GEM-Selektor}, which is then split according to \Cref{sec:mecResults}. A KAN with four hidden layers, 25 neurons per layer, 10th-order activation functions, and 12 grid points is selected, resulting in approximately 56,000 trainable parameters. The network is trained over 100 epochs, with a batch size of 192 and the mean squared error loss function. 
    
    \Cref{fig:ss3Comp} shows the relative errors for all the output variables. The median errors are all below \SI{1e-3}, and no predictions exceed the 10\% error threshold. Even with the added complexity of strontium in the system, the KAN surrogate model provides accurate predictions of the chemical equilibrium.
    
    \begin{table}[h!]
        \centering
        \begin{tabular}{l|c|c|c}
        \textbf{Model} & \textbf{Run time GEMS (s)} & \textbf{Run time KAN (s)} & \textbf{Improvement (\%)} \\
        \hline
        (i) Mech.\ mix & 2.45 & 0.176 & 92.82 \\
        (ii) Binary SS & 1.29 & 0.162 & 87.44 \\
        (iii) Ternary SS & 4.13 & 0.260  & 93.71 \\
        \end{tabular}
        \caption{Comparison of run times between \suite{GEM-Selektor} and the KAN surrogates for the radium uptake case studies. The time required to perform 5,000 equilibria calculations is displayed.}
        \label{table:surrogateRunTime}
    \end{table}
    
    Finally, we investigate the computational efficiency of the surrogate models against the original solver \suite{GEM-Selektor}. \Cref{table:surrogateRunTime} shows the run times for 5,000 equilibrium calculations on an AMD Ryzen Threadripper PRO 5965WX 24-Cores processor. As expected, the computational cost increases with the model complexity, with the ternary SS requiring the most time. Note, however, that the KAN surrogate consistently achieves time reductions exceeding 87\% across all cases. Despite the increased complexity requiring ca.\ 50,000 trainable parameters in case (iii), the surrogate model achieves a $93.7\%$ reduction in computational time---a speedup factor of approximately 16$\times$.
    
    \subsection{Discussion}
    
    Having assessed the performance of KAN surrogate models, we highlight the following findings:
    
    \textbf{Surrogate model accuracy}: KANs achieve a higher prediction accuracy than MLPs across all tested scenarios. KANs exhibit lower median errors on the test set and significantly fewer high-error predictions, see \Cref{fig:mecMixErr}. With enough training data, the KAN surrogate can successfully replace the chemical equilibrium solver. For the most complex ternary SS model, no prediction error exceeds the $10\%$ threshold (see \Cref{fig:ss3Comp}), with median errors consistently below \SI{1e-3}{}.
    
    \textbf{Surrogate model speedup:} Despite longer training times, KANs demonstrate excellent computational efficiency. KAN surrogates achieve 87-93\% reductions in computational time compared to \suite{GEM-Selektor} across the radium uptake test cases (see \Cref{table:surrogateRunTime}).
    
    \textbf{Parameter efficiency:} A notable advantage of KANs is their ability to achieve superior accuracy with fewer trainable parameters compared to MLPs, reducing the memory requirements for training and storage. Both in the cement hydration benchmark and in the radium uptake (i) case, KAN architectures with fewer parameters outperformed larger MLP models. Importantly, all networks were trained on identical datasets, and we did not observe the data inefficiency issues reported in \cite{shukla_comprehensive_2024}.
    
    \textbf{Training stability:} Shukla et al.\ reported loss divergence when training KANs to solve partial differential equations \cite{shukla_comprehensive_2024}. However, in the examined geochemical equilibrium applications, we observed no such instabilities. The transition from MLPs required no modifications or stabilization of the training process.
    
    \textbf{Training performance:} The main limitation of KANs is increased training time. KANs not only train slower than MLPs, but the gap widens significantly with the number of trainable parameters, i.e., the complexity of the spline-based activation functions (see \Cref{table:cementTrainTime}).
    
    In summary, the application of KANs as surrogate models for chemical equilibrium calculations demonstrates clear advantages over classical MLP architectures. The primary limitation---increased training time---represents a one-time investment of at most ten minutes in our test cases. This modest cost is far outweighed by the superior prediction accuracy (up to 62\% error reduction).
    
    This study focused exclusively on one-time predictions of chemical equilibria. Integrating surrogates into reactive transport simulations introduces additional challenges. When chemical reactions are solved at each computational cell and time step, violations of mass conservation can lead to error accumulation over time. To address this challenge, Silva et al.\ enforced charge balance as a penalty term in the loss function \cite{silva_rapid_2025}. An alternative approach would incorporate the Gibbs energy minimization formulation directly into the loss function, analogous to physics-informed NNs for partial differential equations. Investigating these strategies for RTM represents a crucial next step toward deploying KAN-based surrogates.

    \section{Conclusions}
    \label{sec:section5}
    
    In this work, we investigated KANs as surrogate models for computing chemical equilibria. We first tested KANs against classical MLPs on a cement hydration benchmark. Then, we examined three models of radium incorporation into sulfate SSs up to the ternary \ch{(Sr,Ba,Ra)SO4} system. To our knowledge, this represents the first surrogate model for radionuclide co-precipitation, including temperature-dependent systems.
    
    On the cement hydration benchmark, KANs reduce absolute and relative errors by 62\% and 59\%, respectively, compared to MLP architectures. Importantly, KANs achieve this accuracy with fewer trainable parameters, outperforming larger MLP models on identical datasets. This performance extends to more complex systems: for the binary and ternary SS models, KANs maintained median prediction errors near \SI{1e-3}{} with no predictions exceeding the 10\% error threshold. Beyond accuracy, KANs achieve 87-93\% reductions in evaluation time compared to \suite{GEM-Selektor} across all test cases. While KANs require longer training times than MLPs---at most ten minutes in our test cases---this represents a one-time investment that is negligible within the modelling workflow.
    
    The combination of superior accuracy and parameter efficiency make KANs a promising tool to accelerate RTM. Future work should address the extension to time-dependent simulations, including coupling with flow solvers to tackle more complex geochemical scenarios. Additionally, strategies to enforce conservation of mass and prevent error accumulation could further improve the performance of KANs.
    
    This work represents an important step toward efficient real-time uncertainty quantification and high-resolution simulations for complex geochemical systems relevant to nuclear waste management and environmental applications.

    \section*{Open Research Section}

    The employed datasets, as well as the \suite{GEM-Selektor} input files and Python generation scripts, are available at \url{https://doi.org/10.5281/zenodo.18682412}.

	\section*{Declaration of Generative AI and AI-assisted Tools}

    The authors declare that they have used Claude (version Sonnet 4.5) and ChatGPT (version 5.2) to spell-check the manuscript and refine sentences. We emphasize that AI tools have generated no content.

    \section*{Conflict of Interest Declaration}

    The authors declare there are no conflicts of interest for this manuscript.

    \section*{Acknowledgements}

    The authors gratefully acknowledge The Helmholtz Association of German Research Centers (HGF) and the Federal Ministry of Research, Technology and Space (BMFTR), Germany for supporting this work within the frame of the Innovation Pool project “DTN - Digital twins in nuclear waste disposal” and the Helmholtz Research Program “Nuclear Waste Management, Safety and Radiation Research” (NUSAFE).
    The authors would like to thank Ryan Santoso for his constructive criticism on the manuscript.

	\bibliographystyle{elsarticle-num}  
	\bibliography{literature}

@article{steefel_reactive_2005,
	title = {Reactive transport modeling: {An} essential tool and a new research approach for the {Earth} sciences},
	volume = {240},
	copyright = {https://www.elsevier.com/tdm/userlicense/1.0/},
	issn = {0012821X},
	shorttitle = {Reactive transport modeling},
	doi = {10.1016/j.epsl.2005.09.017},
	language = {en},
	number = {3-4},
	journal = {Earth and Planetary Science Letters},
	author = {Steefel, C and Depaolo, D and Lichtner, P},
	year = {2005},
	pages = {539--558},
}

@article{steefel_reactive_2019,
	title = {Reactive {Transport} at the {Crossroads}},
	volume = {85},
	issn = {1529-6466},
	doi = {10.2138/rmg.2019.85.1},
	number = {1},
	journal = {Reviews in Mineralogy and Geochemistry},
	author = {Steefel, Carl I.},
	year = {2019},
	pages = {1--26}
}

@article{kang_pore_2010,
	title = {Pore {Scale} {Modeling} of {Reactive} {Transport} {Involved} in {Geologic} {CO2} {Sequestration}},
	volume = {82},
	issn = {1573-1634},
	doi = {10.1007/s11242-009-9443-9},
	language = {en},
	number = {1},
	journal = {Transport in Porous Media},
	author = {Kang, Qinjun and Lichtner, Peter C. and Viswanathan, Hari S. and Abdel-Fattah, Amr I.},
	year = {2010},
	pages = {197--213}
}

@article{liu_reactive_2019,
	title = {Reactive transport modeling of mineral carbonation in unaltered and altered basalts during {CO2} sequestration},
	volume = {85},
	issn = {1750-5836},
	doi = {10.1016/j.ijggc.2019.04.006},
	journal = {International Journal of Greenhouse Gas Control},
	author = {Liu, Danqing and Agarwal, Ramesh and Li, Yilian and Yang, Sen},
	year = {2019},
	pages = {109--120}
}

@article{yapparova_advanced_2019,
	title = {An advanced reactive transport simulation scheme for hydrothermal systems modelling},
	volume = {78},
	issn = {0375-6505},
	doi = {10.1016/j.geothermics.2018.12.003},
	journal = {Geothermics},
	author = {Yapparova, Alina and Miron, George D. and Kulik, Dmitrii A. and Kosakowski, Georg and Driesner, Thomas},
	year = {2019},
	pages = {138--153}
}

@article{erol_fluid-co2_2022,
	title = {Fluid-{CO2} injection impact in a geothermal reservoir: {Evaluation} with 3-{D} reactive transport modeling},
	volume = {98},
	issn = {0375-6505},
	shorttitle = {Fluid-{CO2} injection impact in a geothermal reservoir},
	doi = {10.1016/j.geothermics.2021.102271},
	journal = {Geothermics},
	author = {Erol, Selçuk and Akın, Taylan and Başer, Ali and Saraçoğlu, {\"O}nder and Akın, Serhat},
	year = {2022},
	pages = {102271}
}

@article{visser_trends_2009,
	series = {Transfer of pollutants in soils, sediments and water systems: {From} small to large scale ({AquaTerra})},
	title = {Trends in pollutant concentrations in relation to time of recharge and reactive transport at the groundwater body scale},
	volume = {369},
	issn = {0022-1694},
	doi = {10.1016/j.jhydrol.2009.02.008},
	number = {3},
	journal = {Journal of Hydrology},
	author = {Visser, Ate and Broers, Hans Peter and Heerdink, Ruth and Bierkens, Marc F. P.},
	year = {2009},
	pages = {427--439}
}

@article{leterme_reactive_2014,
	title = {A reactive transport model for mercury fate in soil—application to different anthropogenic pollution sources},
	volume = {21},
	issn = {1614-7499},
	doi = {10.1007/s11356-014-3135-x},
	language = {en},
	number = {21},
	journal = {Environmental Science and Pollution Research},
	author = {Leterme, Bertrand and Blanc, Philippe and Jacques, Diederik},
	year = {2014},
	pages = {12279--12293}
}

@article{paz-garcia_modeling_2012,
	series = {{EREM} 2011 + {ISEE}'{Cap} 2011 + {EMRS} 2011},
	title = {Modeling of electrokinetic desalination of bricks},
	volume = {86},
	issn = {0013-4686},
	doi = {10.1016/j.electacta.2012.05.132},
	journal = {Electrochimica Acta},
	author = {Paz-García, Juan Manuel and Johannesson, Björn and Ottosen, Lisbeth M. and Alshawabkeh, Akram N. and Ribeiro, Alexandra B. and Rodríguez-Maroto, José Miguel},
	year = {2012},
	pages = {213--222}
}

@article{leal_overview_2017,
	title = {An overview of computational methods for chemical equilibrium and kinetic calculations for geochemical and reactive transport modeling},
	volume = {89},
	copyright = {De Gruyter expressly reserves the right to use all content for commercial text and data mining within the meaning of Section 44b of the German Copyright Act.},
	issn = {1365-3075},
	doi = {10.1515/pac-2016-1107},
	language = {en},
	number = {5},
	journal = {Pure and Applied Chemistry},
	author = {Leal, Allan M. M. and Kulik, Dmitrii A. and Smith, William R. and Saar, Martin O.},
	year = {2017},
	pages = {597--643}
}

@article{leal_accelerating_2020,
	title = {Accelerating {Reactive} {Transport} {Modeling}: {On}-{Demand} {Machine} {Learning} {Algorithm} for {Chemical} {Equilibrium} {Calculations}},
	volume = {133},
	issn = {1573-1634},
	shorttitle = {Accelerating {Reactive} {Transport} {Modeling}},
	doi = {10.1007/s11242-020-01412-1},
	language = {en},
	number = {2},
	journal = {Transport in Porous Media},
	author = {Leal, Allan M. M. and Kyas, Svetlana and Kulik, Dmitrii A. and Saar, Martin O.},
	year = {2020},
	pages = {161--204}
}

@inproceedings{lubke_fast_2025,
	address = {Cham},
	title = {A {Fast} {MPI}-{Based} {Distributed} {Hash}-{Table} as {Surrogate} {Model} for {HPC} {Applications}},
	isbn = {978-3-031-97635-3},
	doi = {10.1007/978-3-031-97635-3_28},
	language = {en},
	booktitle = {Computational {Science} – {ICCS} 2025},
	publisher = {Springer Nature Switzerland},
	author = {Lübke, Max and De Lucia, Marco and Petri, Stefan and Schnor, Bettina},
	editor = {Lees, Michael H. and Cai, Wentong and Cheong, Siew Ann and Su, Yi and Abramson, David and Dongarra, Jack J. and Sloot, Peter M. A.},
	year = {2025},
	pages = {233--240}
}

@article{de_lucia_poet_2021,
	title = {{POET} (v0.1): speedup of many-core parallel reactive transport simulations with fast {DHT} lookups},
	volume = {14},
	issn = {1991-959X},
	shorttitle = {{\textless}span style="" class="text typewriter"{\textgreater}{POET}{\textless}/span{\textgreater} (v0.1)},
	doi = {10.5194/gmd-14-7391-2021},
	language = {English},
	number = {12},
	journal = {Geoscientific Model Development},
	author = {De Lucia, Marco and Kühn, Michael and Lindemann, Alexander and Lübke, Max and Schnor, Bettina},
	year = {2021},
	note = {Publisher: Copernicus GmbH},
	pages = {7391--7409}
}

@article{kyas_accelerated_2022,
	title = {Accelerated reactive transport simulations in heterogeneous porous media using {Reaktoro} and {Firedrake}},
	volume = {26},
	issn = {1573-1499},
	doi = {10.1007/s10596-021-10126-2},
	language = {en},
	number = {2},
	journal = {Computational Geosciences},
	author = {Kyas, Svetlana and Volpatto, Diego and Saar, Martin O. and Leal, Allan M. M.},
	year = {2022},
	pages = {295--327}
}

@article{laloy_speeding_2022,
	title = {Speeding {Up} {Reactive} {Transport} {Simulations} in {Cement} {Systems} by {Surrogate} {Geochemical} {Modeling}: {Deep} {Neural} {Networks} and k-{Nearest} {Neighbors}},
	volume = {143},
	issn = {1573-1634},
	shorttitle = {Speeding {Up} {Reactive} {Transport} {Simulations} in {Cement} {Systems} by {Surrogate} {Geochemical} {Modeling}},
	doi = {10.1007/s11242-022-01779-3},
	language = {en},
	number = {2},
	journal = {Transport in Porous Media},
	author = {Laloy, Eric and Jacques, Diederik},
	year = {2022},
	pages = {433--462}
}

@article{laloy_emulation_2019,
	title = {Emulation of {CPU}-demanding reactive transport models: a comparison of {Gaussian} processes, polynomial chaos expansion, and deep neural networks},
	volume = {23},
	issn = {1573-1499},
	shorttitle = {Emulation of {CPU}-demanding reactive transport models},
	doi = {10.1007/s10596-019-09875-y},
	language = {en},
	number = {5},
	journal = {Computational Geosciences},
	author = {Laloy, Eric and Jacques, Diederik},
	year = {2019},
	pages = {1193--1215}
}

@article{demirer_improving_2023,
	title = {Improving the {Performance} of {Reactive} {Transport} {Simulations} {Using} {Artificial} {Neural} {Networks}},
	volume = {149},
	issn = {1573-1634},
	doi = {10.1007/s11242-022-01856-7},
	language = {en},
	number = {1},
	journal = {Transport in Porous Media},
	author = {Demirer, Ersan and Coene, Emilie and Iraola, Aitor and Nardi, Albert and Abarca, Elena and Idiart, Andrés and de Paola, Giorgio and Rodríguez-Morillas, Noelia},
	year = {2023},
	pages = {271--297}
}

@misc{silva_rapid_2025,
	title = {Rapid modelling of reactive transport in porous media using machine learning: limitations and solutions},
	shorttitle = {Rapid modelling of reactive transport in porous media using machine learning},
	doi = {10.48550/arXiv.2405.14548},
	publisher = {arXiv},
	author = {Silva, Vinicius L. S. and Regnier, Geraldine and Salinas, Pablo and Heaney, Claire E. and Jackson, Matthew D. and Pain, Christopher C.},
	year = {2025},
	note = {arXiv:2405.14548 [cs]}
}

@article{prasianakis_geochemistry_2025,
	title = {Geochemistry and machine learning: methods and benchmarking},
	volume = {84},
	issn = {1866-6299},
	shorttitle = {Geochemistry and machine learning},
	doi = {10.1007/s12665-024-12066-3},
	language = {en},
	number = {5},
	journal = {Environmental Earth Sciences},
	author = {Prasianakis, N. I. and Laloy, E. and Jacques, D. and Meeussen, J. C. L. and Miron, G. D. and Kulik, D. A. and Idiart, A. and Demirer, E. and Coene, E. and Cochepin, B. and Leconte, M. and Savino, M. E. and Samper-Pilar, J. and De Lucia, M. and Churakov, S. V. and Kolditz, O. and Yang, C. and Samper, J. and Claret, F.},
	year = {2025},
	pages = {121}
}

@article{guo_physics-informed_2025,
	title = {Physics-informed {Kolmogorov}–{Arnold} network with {Chebyshev} polynomials for fluid mechanics},
	volume = {37},
	issn = {1070-6631},
	doi = {10.1063/5.0284999},
	number = {9},
	journal = {Physics of Fluids},
	author = {Guo, Chunyu and Sun, Lucheng and Li, Shilong and Yuan, Zelong and Wang, Chao},
	year = {2025},
	pages = {095120}
}

@misc{xu_kolmogorov-arnold_2024,
	title = {Kolmogorov-{Arnold} {Networks} for {Time} {Series}: {Bridging} {Predictive} {Power} and {Interpretability}},
	shorttitle = {Kolmogorov-{Arnold} {Networks} for {Time} {Series}},
	doi = {10.48550/arXiv.2406.02496},
	publisher = {arXiv},
	author = {Xu, Kunpeng and Chen, Lifei and Wang, Shengrui},
	year = {2024},
	note = {arXiv:2406.02496 [cs]}
}

@article{li_kolmogorovarnold_2025,
	title = {Kolmogorov–{Arnold} graph neural networks for molecular property prediction},
	volume = {7},
	copyright = {2025 The Author(s)},
	issn = {2522-5839},
	doi = {10.1038/s42256-025-01087-7},
	language = {en},
	number = {8},
	journal = {Nature Machine Intelligence},
	author = {Li, Longlong and Zhang, Yipeng and Wang, Guanghui and Xia, Kelin},
	year = {2025},
	note = {Publisher: Nature Publishing Group},
	pages = {1346--1354}
}

@misc{liu_kan_2025,
	title = {{KAN}: {Kolmogorov}-{Arnold} {Networks}},
	shorttitle = {{KAN}},
	doi = {10.48550/arXiv.2404.19756},
	publisher = {arXiv},
	author = {Liu, Ziming and Wang, Yixuan and Vaidya, Sachin and Ruehle, Fabian and Halverson, James and Soljačić, Marin and Hou, Thomas Y. and Tegmark, Max},
	year = {2025},
	note = {arXiv:2404.19756 [cs]}
}

@misc{kiamari_gkan_2024,
	title = {{GKAN}: {Graph} {Kolmogorov}-{Arnold} {Networks}},
	shorttitle = {{GKAN}},
	doi = {10.48550/arXiv.2406.06470},
	publisher = {arXiv},
	author = {Kiamari, Mehrdad and Kiamari, Mohammad and Krishnamachari, Bhaskar},
	year = {2024},
	note = {arXiv:2406.06470 [cs]}
}

@misc{bodner_convolutional_2025,
	title = {Convolutional {Kolmogorov}-{Arnold} {Networks}},
	doi = {10.48550/arXiv.2406.13155},
	publisher = {arXiv},
	author = {Bodner, Alexander Dylan and Tepsich, Antonio Santiago and Spolski, Jack Natan and Pourteau, Santiago},
	year = {2025},
	note = {arXiv:2406.13155 [cs]},
}

@article{shukla_comprehensive_2024,
	title = {A comprehensive and {FAIR} comparison between {MLP} and {KAN} representations for differential equations and operator networks},
	volume = {431},
	issn = {0045-7825},
	doi = {10.1016/j.cma.2024.117290},
	journal = {Computer Methods in Applied Mechanics and Engineering},
	author = {Shukla, Khemraj and Toscano, Juan Diego and Wang, Zhicheng and Zou, Zongren and Karniadakis, George Em},
	year = {2024},
	pages = {117290}
}

@misc{hou_kolmogorov-arnold_2025,
	title = {Kolmogorov-{Arnold} {Networks}: {A} {Critical} {Assessment} of {Claims}, {Performance}, and {Practical} {Viability}},
	shorttitle = {Kolmogorov-{Arnold} {Networks}},
	doi = {10.48550/arXiv.2407.11075},
	language = {en},
	publisher = {arXiv},
	author = {Hou, Yuntian and Ji, Tianrui and Zhang, Di and Stefanidis, Angelos},
	year = {2025},
	note = {arXiv:2407.11075 [cs]}
}

@article{toscano_pinns_2025,
	title = {From {PINNs} to {PIKANs}: recent advances in physics-informed machine learning},
	volume = {1},
	issn = {3005-1436},
	shorttitle = {From {PINNs} to {PIKANs}},
	doi = {10.1007/s44379-025-00015-1},
	language = {en},
	number = {1},
	journal = {Machine Learning for Computational Science and Engineering},
	author = {Toscano, Juan Diego and Oommen, Vivek and Varghese, Alan John and Zou, Zongren and Ahmadi Daryakenari, Nazanin and Wu, Chenxi and Karniadakis, George Em},
	year = {2025},
	pages = {15}
}

@article{jacob_spikans_2025,
	title = {{SPIKANs}: separable physics-informed {Kolmogorov}–{Arnold} networks},
	volume = {6},
	issn = {2632-2153},
	shorttitle = {{SPIKANs}},
	doi = {10.1088/2632-2153/ae05af},
	language = {en},
	number = {3},
	journal = {Machine Learning: Science and Technology},
	author = {Jacob, Bruno and Howard, Amanda A and Stinis, Panos},
	year = {2025},
	note = {Publisher: IOP Publishing},
	pages = {035060}
}

@article{wagner_gems_2012,
	title = {{GEM}-{SELEKTOR} {GEOCHEMICAL} {MODELING} {PACKAGE}: {TSolMod} {LIBRARY} {AND} {DATA} {INTERFACE} {FOR} {MULTICOMPONENT} {PHASE} {MODELS}},
	volume = {50},
	issn = {0008-4476},
	shorttitle = {{GEM}-{SELEKTOR} {GEOCHEMICAL} {MODELING} {PACKAGE}},
	doi = {10.3749/canmin.50.5.1173},
	number = {5},
	journal = {The Canadian Mineralogist},
	author = {Wagner, Thomas and Kulik, Dmitrii A. and Hingerl, Ferdinand F. and Dmytrieva, Svitlana V.},
	year = {2012},
	pages = {1173--1195}
}

@article{kulik_gem-selektor_2013,
	title = {{GEM}-{Selektor} geochemical modeling package: revised algorithm and {GEMS3K} numerical kernel for coupled simulation codes},
	volume = {17},
	issn = {1573-1499},
	shorttitle = {{GEM}-{Selektor} geochemical modeling package},
	doi = {10.1007/s10596-012-9310-6},
	language = {en},
	number = {1},
	journal = {Computational Geosciences},
	author = {Kulik, Dmitrii A. and Wagner, Thomas and Dmytrieva, Svitlana V. and Kosakowski, Georg and Hingerl, Ferdinand F. and Chudnenko, Konstantin V. and Berner, Urs R.},
	year = {2013},
	pages = {1--24}
}

@incollection{williams_novel_2021,
	title = {Novel {Tool} for {Selecting} {Surrogate} {Modeling} {Techniques} for {Surface} {Approximation}},
	volume = {50},
	copyright = {https://www.elsevier.com/tdm/userlicense/1.0/},
	isbn = {978-0-323-88506-5},
	language = {en},
	booktitle = {Computer {Aided} {Chemical} {Engineering}},
	publisher = {Elsevier},
	author = {Williams, Bianca and Cremaschi, Selen},
	year = {2021},
	doi = {10.1016/B978-0-323-88506-5.50071-1},
	pages = {451--456}
}

@misc{hounmenou_formalism_2021,
	title = {A {Formalism} of the {General} {Mathematical} {Expression} of {Multilayer} {Perceptron} {Neural} {Networks}},
	doi = {10.20944/preprints202105.0412.v1},
	language = {en},
	publisher = {MATHEMATICS \& COMPUTER SCIENCE},
	author = {Hounmenou, Castro Gbememali and Gneyou, Kossi Essona and Glele Kakaï, Romain Lucas},
	year = {2021}
}

@misc{Falcon_PyTorch_Lightning_2019,
    author = {Falcon, William and {The PyTorch Lightning team}},
    doi = {10.5281/zenodo.3828935},
    license = {Apache-2.0},
    title = {{PyTorch Lightning}},
    url = {https://github.com/Lightning-AI/lightning},
    version = {1.4},
    year = {2019}
}

@inproceedings{akiba_optuna_2019,
	address = {New York, NY, USA},
	series = {{KDD} '19},
	title = {Optuna: {A} {Next}-generation {Hyperparameter} {Optimization} {Framework}},
	isbn = {978-1-4503-6201-6},
	doi = {10.1145/3292500.3330701},
	booktitle = {Proceedings of the 25th {ACM} {SIGKDD} {International} {Conference} on {Knowledge} {Discovery} \& {Data} {Mining}},
	publisher = {Association for Computing Machinery},
	author = {Akiba, Takuya and Sano, Shotaro and Yanase, Toshihiko and Ohta, Takeru and Koyama, Masanori},
	year = {2019},
	pages = {2623--2631},
}

@article{chan_deep_2023,
	title = {Deep neural networks in the cloud: {Review}, applications, challenges and research directions},
	volume = {545},
	issn = {0925-2312},
	doi = {10.1016/j.neucom.2023.126327},
	journal = {Neurocomputing},
	author = {Chan, Kit Yan and Abu-Salih, Bilal and Qaddoura, Raneem and Al-Zoubi, Ala’ M. and Palade, Vasile and Pham, Duc-Son and Ser, Javier Del and Muhammad, Khan},
	year = {2023},
	pages = {126327}
}

@book{prautzsch_bezier_2002,
	address = {Berlin, Heidelberg},
	series = {Mathematics and {Visualization}},
	title = {Bézier and {B}-{Spline} {Techniques}},
	copyright = {http://www.springer.com/tdm},
	isbn = {978-3-642-07842-2 978-3-662-04919-8},	publisher = {Springer},
	author = {Prautzsch, Hartmut and Boehm, Wolfgang and Paluszny, Marco},
	year = {2002},
	doi = {10.1007/978-3-662-04919-8}
}

@misc{Blealtan_efficient_kan_2024,
    author = {Blealtan, Cao and Dash, Akaash},
    license = {MIT License},
    title = {{efficient-kan}},
    url = {https://github.com/Blealtan/efficient-kan},
    version = {0.1.0},
    year = {2024}
}

@misc{prasianakis_repo_2025,
	title = {Geochemistry and {Machine} {Learning}: {Methods} and {Benchmarking}},
	shorttitle = {Geochemistry and {Machine} {Learning}},
	doi = {10.5281/zenodo.14904784},
	publisher = {Zenodo},
	author = {Prasianakis, N.I. and Laloy, E. and Jacques, D. and Meeussen, J.C.L. and Tournassat, C. and Miron, G.D. and Kulik, D.A. and Idiart, A. and Demirer, E. and Coene, E. and Cochepin, B. and Leconte, M. and Savino, M.E. and Samper II, J. and De Lucia, M. and Churakov, S.V. and Kolditz, O. and Yang, C. and Samper, J. and Claret, F.},
	year = {2025},
}

@article{peng_machine_2025,
	title = {Machine {Learning}-{Enhanced} {Modeling} of {Calcium} {Carbonate} {Nucleation} in {Porous} {Media} {Under} {Counter}-{Diffusion} {Conditions}},
	volume = {61},
	issn = {1944-7973},
	doi = {10.1029/2025WR040484},
	language = {en},
	number = {11},
	journal = {Water Resources Research},
	author = {Peng, Haonan and Rajyaguru, Ashish and Curti, Enzo and Grolimund, Daniel and Churakov, Sergey V. and Prasianakis, Nikolaos I.},
	year = {2025},
	pages = {e2025WR040484}
}

@article{SAMPER2025106286,
    title = {Global sensitivity analysis of reactive transport modelling for the geochemical evolution of a high-level radioactive waste repository},
	volume = {180},
	issn = {0883-2927},
	doi = {10.1016/j.apgeochem.2025.106286},
	urldate = {2026-01-20},
	journal = {Applied Geochemistry},
	author = {Samper, J. and López-Vázquez, C. and Pisani, B. and Mon, A. and Samper-Pilar, A. C. and Samper-Pilar, F. J.},
	year = {2025},
	pages = {106286}
}

@article{MONTENEGRO2023107018,
    title = {A non-isothermal reactive transport model of the long-term geochemical evolution at the disposal cell scale in a HLW repository in granite},
    journal = {Applied Clay Science},
    volume = {242},
    pages = {107018},
    year = {2023},
    issn = {0169-1317},
    doi = {https://doi.org/10.1016/j.clay.2023.107018},
    author = {Luis Montenegro and Javier Samper and Alba Mon and Laurent {De Windt} and Aurora-Core Samper and Enrique García}
}

@article{Kolditz2023Digitalisation,
	title = {Digitalisation for nuclear waste management: predisposal and disposal},
	volume = {82},
	issn = {1866-6299},
	shorttitle = {Digitalisation for nuclear waste management},
	doi = {10.1007/s12665-022-10675-4},
	language = {en},
	number = {1},
	journal = {Environmental Earth Sciences},
	author = {Kolditz, Olaf and Jacques, Diederik and Claret, Francis and Bertrand, Johan and Churakov, Sergey V. and Debayle, Christophe and Diaconu, Daniela and Fuzik, Kateryna and Garcia, David and Graebling, Nico and Grambow, Bernd and Holt, Erika and Idiart, Andrés and Leira, Petter and Montoya, Vanessa and Niederleithinger, Ernst and Olin, Markus and Pfingsten, Wilfried and Prasianakis, Nikolaos I. and Rink, Karsten and Samper, Javier and Szöke, István and Szöke, Réka and Theodon, Louise and Wendling, Jacques},
	year = {2023},
	pages = {42}
}

@article{Brandt20201,
	author = {Brandt, Felix and Klinkenberg, Martina and Poonoosamy, Jenna and Bosbach, Dirk},
	title = {Recrystallization and {Uptake} of {$^{226}$Ra} into {Ba}-{Rich} \ch{({Ba},{Sr}){SO4}} {Solid} {Solutions}},
	year = {2020},
	journal = {Minerals},
	volume = {10},
	number = {9},
	pages = {1 – 28},
	doi = {10.3390/min10090812}
}

@article{Brandt2018,
	author = {Brandt, Felix and Klinkenberg, Martina and Poonoosamy, Jenna and Weber, Juliane and Bosbach, Dirk},
	title = {The {Effect} of {Ionic} {Strength} and {Sr}$_{\text{aq}}$ upon the {Uptake} of {Ra} during the {Recrystallization} of {Barite}},
	year = {2018},
	journal = {Minerals},
	volume = {8},
	number = {11},
	doi = {10.3390/min8110502}
}

@article{Klinkenberg20181,
	author = {Klinkenberg, Martina and Weber, Juliane and Barthel, Juri and Vinograd, Victor and Poonoosamy, Jenna and Kruth, Maximilian and Bosbach, Dirk and Brandt, Felix},
	title = {The solid solution–aqueous solution system \ch{(Sr,Ba,Ra)SO4} + \ch{H2O}: {A} combined experimental and theoretical study of phase equilibria at \ch{Sr}-rich compositions},
	year = {2018},
	journal = {Chemical Geology},
	volume = {497},
	pages = {1 – 17},
	doi = {10.1016/j.chemgeo.2018.08.009}
}

@article{Vinograd2018190,
	author = {Vinograd, V.L. and Kulik, D.A. and Brandt, F. and Klinkenberg, M. and Weber, J. and Winkler, B. and Bosbach, D.},
	title = {Thermodynamics of the solid solution - {Aqueous} solution system \ch{(Ba,Sr,Ra)SO4} + \ch{H2O}: {II}. {Radium} retention in barite-type minerals at elevated temperatures},
	year = {2018},
	journal = {Applied Geochemistry},
	volume = {93},
	pages = {190 – 208},
	doi = {10.1016/j.apgeochem.2017.10.019}
}

@article{Vinograd201859,
	author = {Vinograd, V.L. and Kulik, D.A. and Brandt, F. and Klinkenberg, M. and Weber, J. and Winkler, B. and Bosbach, D.},
	title = {Thermodynamics of the solid solution - {Aqueous} solution system \ch{(Ba,Sr,Ra)SO4} + \ch{H2O}: {I}. {The} effect of strontium content on radium uptake by barite},
	year = {2018},
	journal = {Applied Geochemistry},
	volume = {89},
	pages = {59 – 74},
	doi = {10.1016/j.apgeochem.2017.11.009}
}

@article{Weber2017722,
	author = {Weber, Juliane and Barthel, Juri and Klinkenberg, Martina and Bosbach, Dirk and Kruth, Maximilian and Brandt, Felix},
	title = {Retention of $^{226}$\ch{Ra} by barite: {The} role of internal porosity},
	year = {2017},
	journal = {Chemical Geology},
	volume = {466},
	pages = {722 – 732},
	doi = {10.1016/j.chemgeo.2017.07.021}
}

@article{Brandt20151,
	author = {Brandt, F. and Curti, E. and Klinkenberg, M. and Rozov, K. and Bosbach, D.},
	title = {Replacement of barite by a \ch{(Ba,Ra)SO4} solid solution at close-to-equilibrium conditions: {A} combined experimental and theoretical study},
	year = {2015},
	journal = {Geochimica et Cosmochimica Acta},
	volume = {155},
	pages = {1 – 15},
	doi = {10.1016/j.gca.2015.01.016}
}

@article{Klinkenberg20146620,
	author = {Klinkenberg, Martina and Brandt, Felix and Breuer, Uwe and Bosbach, Dirk},
	title = {Uptake of \ch{Ra} during the {Recrystallization} of {Barite}: {A} {Microscopic} and {Time} of {Flight}-{Secondary} {Ion} {Mass} {Spectrometry} {Study}},
    year = {2014},
	journal = {Environmental Science and Technology},
	volume = {48},
	number = {12},
	pages = {6620 – 6627},
	doi = {10.1021/es405502e}
}

@article{Vinograd2013398,
	author = {Vinograd, V.L. and Brandt, F. and Rozov, K. and Klinkenberg, M. and Refson, K. and Winkler, B. and Bosbach, D.},
	title = {Solid-aqueous equilibrium in the \ch{BaSO4-RaSO4-H2O} system: {First}-principles calculations and a thermodynamic assessment},
	year = {2013},
	journal = {Geochimica et Cosmochimica Acta},
	volume = {122},
	pages = {398 – 417},
	doi = {10.1016/j.gca.2013.08.028}
}

@article{Poonoosamy2024Radiochemical,
  author  = {Poonoosamy, J. and Kaspor, A. and Schreinemachers, C. and Bosbach, D. and Cheong, O. and Kowalski, P. M. and Obaied, A.},
  title   = {A radiochemical lab-on-a-chip paired with computer vision to unlock the crystallization kinetics of \ch{(Ba,Ra)SO4}},
  journal = {Scientific Reports},
  volume  = {14},
  pages   = {9502},
  year    = {2024},
  doi     = {10.1038/s41598-024-59888-6}
}

@book{hummel2002nagra,
    title = {Nagra/{PSI} {Chemical} {Thermodynamic} {Data} {Base} 01/01},
	isbn = {978-1-58112-620-4},
	language = {en},
	publisher = {Universal-Publishers},
	author = {Hummel, Wolfgang and Berner, Urs and Curti, Enzo and Pearson, F. J. and Thoenen, Tres},
	year = {2002}
}

@article{thoenen2014psi,
    title={{The} {PSI/Nagra} {Chemical} {Thermodynamic} {Database} 12/07},
    author={Thoenen, Tres and Hummel, Wolfgang and Berner, Urs and Curti, Enzo},
    year={2014},
    issn={1019-0643}
}

@article{Helgeson19811249,
	author = {Helgeson, H.C. and Kirkham, D.H. and Flowers, G.C.},
	title = {Theoretical prediction of the thermodynamic behavior of aqueous electrolytes by high pressures and temperatures; {IV}, {Calculation} of activity coefficients, osmotic coefficients, and apparent molal and standard and relative partial molal properties to 600 degrees {C} and 5kb},
    year = {1981},
	journal = {American Journal of Science},
	volume = {281},
	number = {10},
	pages = {1249 – 1516},
	doi = {10.2475/ajs.281.10.1249}
}

@article{Johnson1992899,
	author = {Johnson, James W. and Oelkers, Eric H. and Helgeson, Harold C.},
	title = {{SUPCRT92}: {A} software package for calculating the standard molal thermodynamic properties of minerals, gases, aqueous species, and reactions from 1 to 5000 bar and 0 to 1000°{C}},
    year = {1992},
	journal = {Computers and Geosciences},
	volume = {18},
	number = {7},
	pages = {899 – 947},
	doi = {10.1016/0098-3004(92)90029-Q}
}

@article{sobol_distribution_1967,
	title = {On the distribution of points in a cube and the approximate evaluation of integrals},
	volume = {7},
	issn = {0041-5553},
	doi = {10.1016/0041-5553(67)90144-9},
	number = {4},
	journal = {USSR Computational Mathematics and Mathematical Physics},
	author = {Sobol', I. M},
	year = {1967},
	pages = {86--112}
}

@article{wang_contrasting_2026,
	title = {Contrasting coprecipitation and recrystallization mechanisms for {Ra} immobilization via ({Ba},{Ra}){SO$_4$} solid solution formation in fractured crystalline rocks: {Insights} from {3D} reactive transport modeling},
	issn = {0016-7037},
	doi = {10.1016/j.gca.2026.01.045},
	journal = {Geochimica et Cosmochimica Acta},
	author = {Wang, Yumeng and Alt-Epping, Peter and Deissmann, Guido and Yang, Yuankai and Hu, Jun and Bosbach, Dirk and Poonoosamy, Jenna},
	year = {2026}
}

@techreport{skb_report_2022,
    author      = {SKB},
    title       = {{RD\&D Programme 2022. Programme for research, development and demonstration of methods for the management and disposal of nuclear waste}},
    institution = {Svensk Kärnbränslehantering AB},
    year        = {2022},
    number      = {TR-22-11}
}

@article{zhang_co-precipitation_2014,
	title = {Co-precipitation of {Radium} with {Barium} and {Strontium} {Sulfate} and {Its} {Impact} on the {Fate} of {Radium} during {Treatment} of {Produced} {Water} from {Unconventional} {Gas} {Extraction}},
	volume = {48},
	issn = {0013-936X},
	doi = {10.1021/es405168b},
	number = {8},
	urldate = {2026-02-11},
	journal = {Environmental Science \& Technology},
	publisher = {American Chemical Society},
	author = {Zhang, Tieyuan and Gregory, Kelvin and Hammack, Richard W. and Vidic, Radisav D.},
	year = {2014},
	pages = {4596--4603},
}

@article{claret_eurad_2024,
	title = {{EURAD} state-of-the-art report: development and improvement of numerical methods and tools for modeling coupled processes in the field of nuclear waste disposal},
	volume = {3},
	issn = {2813-3412},
	shorttitle = {{EURAD} state-of-the-art report},
	doi = {10.3389/fnuen.2024.1437714},
	journal = {Frontiers in Nuclear Engineering},
	publisher = {Frontiers},
	author = {Claret, F. and Prasianakis, N. I. and Baksay, A. and Lukin, D. and Pepin, G. and Ahusborde, E. and Amaziane, B. and Bátor, G. and Becker, D. and Bednár, A. and Béreš, M. and Bérešová, S. and Böthi, Z. and Brendler, V. and Brenner, K. and Březina, J. and Chave, F. and Churakov, S. V. and Hokr, M. and Horák, D. and Jacques, D. and Jankovský, F. and Kazymyrenko, C. and Koudelka, T. and Kovács, T. and Krejčí, T. and Kruis, J. and Laloy, E. and Landa, J. and Ligurský, T. and Lipping, T. and López-Vázquez, C. and Masson, R. and Meeussen, J. C. L. and Mollaali, M. and Mon, A. and Montenegro, L. and Pisani, B. and Poonoosamy, J. and Pospiech, S. I. and Saâdi, Z. and Samper, J. and Samper-Pilar, A.-C. and Scaringi, G. and Sysala, S. and Yoshioka, K. and Yang, Y. and Zuna, M. and Kolditz, O.},
	year = {2024}
}
	
\end{document}